\documentclass[lettersize,journal]{IEEEtran}
\usepackage{array}
\usepackage{textcomp}
\usepackage{stfloats}
\usepackage{verbatim}
\usepackage{enumitem}
\usepackage{ulem}
\usepackage{url}
\usepackage{hyperref}
\usepackage{cite}
\usepackage{orcidlink}

\usepackage{amsthm,amssymb,amsmath,amsfonts}
\usepackage{mathtools}

\usepackage{amsmath,amsfonts,bm}

\def\1{\bm{1}}

\def\vv{{\bm{v}}}

\def\vx{{\bm{x}}}

\def\mA{{\bm{A}}}
\def\mB{{\bm{B}}}
\def\mC{{\bm{C}}}
\def\mD{{\bm{D}}}

\def\mG{{\bm{G}}}
\def\mH{{\bm{H}}}
\def\mI{{\bm{I}}}

\def\mS{{\bm{S}}}

\def\mU{{\bm{U}}}

\def\mW{{\bm{W}}}
\def\mX{{\bm{X}}}
\def\mY{{\bm{Y}}}
\def\mZ{{\bm{Z}}}

\DeclareMathAlphabet{\mathsfit}{\encodingdefault}{\sfdefault}{m}{sl}
\SetMathAlphabet{\mathsfit}{bold}{\encodingdefault}{\sfdefault}{bx}{n}

\usepackage{graphicx}
\usepackage{caption}
\usepackage{subcaption}
\usepackage{algpseudocode,algorithm}
\usepackage{multirow}
\usepackage{booktabs}

\usepackage{algorithm}
\usepackage{algpseudocode,algorithm}

\usepackage{tikz}
\usepackage{xcolor}

\theoremstyle{remark} 
\newtheorem{remark}{Remark}

\theoremstyle{definition}

\theoremstyle{definition}

\begin{document}

\title{IGNN-Solver: A Graph Neural Solver for \\ Implicit Graph Neural Networks}


\author{
	Junchao Lin~\orcidlink{0000-0002-6173-4601},
	Zenan Ling~\orcidlink{0009-0001-3975-4403},~\IEEEmembership{Member,~IEEE},
	Zhanbo Feng~\orcidlink{0009-0002-4529-7357},~\IEEEmembership{Student Member,~IEEE}, 
        Jingwen Xu~\orcidlink{0009-0002-3934-3265},
        Minxuan Liao~\orcidlink{0009-0002-6777-2702},
        Feng Zhou~\orcidlink{0000-0003-0842-306X},
        Tianqi Hou~\orcidlink{0000-0003-1940-4833},
        Zhenyu Liao~\orcidlink{0000-0002-1915-8559},~\IEEEmembership{Member,~IEEE},
	and
	Robert C. Qiu~\orcidlink{0000-0002-1154-1836},~\IEEEmembership{Fellow,~IEEE}
    \thanks{Manuscript created on February 17, 2025. The work of Zenan Ling is supported in part by the National Natural Science Foundation of China (via NSFC-62406119), the Natural Science Foundation of Hubei Province (2024AFB074), and the Guangdong Provincial Key Laboratory of Mathematical Foundations for Artificial Intelligence (2023B1212010001). The work of Zhenyu Liao is supported in part by the National Natural Science Foundation of China (via  NSFC-62206101). The work of Robert Caiming Qiu is supported in part by the National Natural Science Foundation of China (via NSFC-12141107) and in part by the Key Research and Development Program of Guangxi (GuiKe-AB21196034).
    \textit{(Corresponding author: Zenan Ling.)}}
	\thanks{Junchao Lin, Zenan Ling, Minxuan Liao, Zhenyu Liao, and Robert C. Qiu are with the School of EIC, Huazhong University of Science and Technology.}
	\thanks{Zhanbo Feng is with the Department of Computer Science and Engineering, Shanghai Jiao Tong University.}
	\thanks{Feng Zhou is with the Center for Applied Statistics and School of Statistics, Renmin University of China.}
    \thanks{Tianqi Hou is with the Lagrange Mathematics and Computation Research Center of Huawei.}
	\thanks{Jingwen Xu is with the School of Science, Wuhan University of Technology.}

}

\markboth{Journal of \LaTeX\ Class Files,~Vol.~14, No.~8, August~2021}%
{Shell \MakeLowercase{\textit{et al.}}: A Sample Article Using IEEEtran.cls for IEEE Journals}


\maketitle

\begin{abstract}
    Implicit graph neural networks (IGNNs), which exhibit strong expressive power with a single layer, have recently demonstrated remarkable performance in capturing long-range dependencies (LRD) in underlying graphs while effectively mitigating the over-smoothing problem. However, IGNNs rely on computationally expensive fixed-point iterations, which lead to significant speed and scalability limitations, hindering their application to large-scale graphs. To achieve fast fixed-point solving for IGNNs, we propose a novel graph neural solver, IGNN-Solver, which leverages the generalized Anderson Acceleration method, parameterized by a tiny GNN, and learns iterative updates as a graph-dependent temporal process. 
    To improve effectiveness on large-scale graph tasks, we further integrate sparsification and storage compression methods, specifically tailored for the IGNN-Solver, into its design.
    Extensive experiments demonstrate that the IGNN-Solver significantly accelerates inference on both small- and large-scale tasks, achieving a $1.5\times$ to $8\times$ speedup without sacrificing accuracy. 
    This advantage becomes more pronounced as the graph scale grows, facilitating its large-scale deployment in real-world applications. The code to reproduce our results is available at \url{https://github.com/landrarwolf/IGNN-Solver}.
\end{abstract}


\section{Introduction}
\label{sec1}
\IEEEPARstart{I}{mplicit} graph neural networks (IGNNs)~\cite{IGNN, EIGNN, MIGNN} have emerged as a significant advancement in graph learning frameworks. Unlike traditional graph neural networks (GNNs) that stack multiple explicit layers, IGNNs utilize a single implicit layer formulated as a fixed-point equation. The solution to this fixed-point equation, known as the equilibrium, is equivalent to the output obtained by iterating an explicit layer infinitely. This allows the implicit layer to access infinite hops of neighbors, providing IGNNs with global receptive fields within just one layer~\cite{chen2022efficient}. As a result, IGNNs effectively address the long-standing issue of over-smoothing in conventional explicit GNNs and capture of long-range dependencies in graph-structured data~\cite{li2018deeper, alon2020bottleneck, zhang2023comprehensive}.

However, existing IGNNs suffer from slow speeds and have difficulty scaling to large-scale graphs~\cite{liu2024scalable, geng2021training, yang2021implicit, nastorg2023implicit}. This is primarily because IGNNs derive features by solving for fixed points, demanding substantial computational resources.  For example, even on the Citeseer dataset~\cite{GCN} classification task with a small-scale graph, IGNNs require more than $20$ forward iterative computations to nearly converge to a fixed point ~\cite{IGNN}. The computational overhead of solving fixed points is amplified by the task scale, resulting in notably slow inference speeds compared to explicit GNNs.  This substantial drawback poses challenges for IGNNs in generalizing to large-scale graphs in practical scenarios.

In response to this challenge, we propose a novel graph neural solver for IGNNs, termed IGNN-Solver. It takes the IGNN layer as input, which includes the graph information matrix and layer parameters, and outputs the solution to the corresponding fixed-point equation.
Unlike conventional solvers, which compute output features through iterative forward passes in a potentially large GNN (relying on root-finding algorithms like Broyden's method~\cite{anderson1965iterative}), the IGNN-Solver predicts the next iteration step using a \textit{tiny} graph network.
Consequently, the proposed IGNN-Solver remarkably accelerates the model's inference speed without compromising accuracy and with only a slight increase in training overhead. This advantage becomes increasingly pronounced as the scale of the graph grows, making it particularly beneficial for deploying IGNNs in large-scale graph tasks.

Our IGNN-Solver comprises two components. First, we introduce a learnable initializer that estimates an optimal initial point for the optimization process. Second, we propose a generalized version of Anderson Acceleration (AA) \cite{anderson1965iterative}, employing a tiny graph network to model the iterative updates as a sequence of graph-dependent steps. Compared to the solvers proposed for conventional Implicit Neural Networks (INNs)~\cite{DEQ, MDEQ, bai2021neural}, we introduce novel improvements: learning solver parameters through a GNN-based method. This approach circumvents the potential loss of graph information, thereby improving the model's performance. Moreover, IGNN-Solver has significantly fewer parameters compared to IGNN, and its training is independent of the IGNN's inference. Consequently, the training of IGNN-Solver proceeds rapidly without sacrificing generalization. 

Moreover, to address the challenges posed by large-scale graphs, we incorporate both graph sparsification and storage compression in the solver design. Specifically, to prevent the network from becoming excessively large and deviating from the initial goal of using a compact network for prediction~\cite{RPI-Graph, zheng2020robust}, we utilize graph sparsification for a lighter graph. Additionally, directly mapping data from high to extremely low dimensions is not appropriate~\cite{koppen2000curse}, thereby we employ storage compression technique to mitigate this issue.


In our experiments, we apply IGNN-Solver to nine real-world datasets from diverse domains and scales, including four large-scale datasets: Amazon-all~\cite{Amazonall}, Reddit~\cite{graphsage}, ogbn-arxiv~\cite{OGB} and ogbn-products~\cite{OGB}. Our results demonstrate that the IGNN-Solver achieves higher accuracy with reduced inference time, showing a $1.5\times$ to $8\times$ speedup, and incurs minimal additional training overhead, constituting only $1\%$ of the IGNN training time.

Our main contributions are summarized as follows:
\begin{itemize}
    \item We introduce IGNN-Solver, a method designed to predict the next fixed-point iteration of the IGNN layer. This innovative approach mitigates the need for extensive iterations, common in vanilla IGNNs, thereby substantially accelerating inference while preserving accuracy and minimizing parameter consumption.
    \item Compared to conventional solvers proposed for INNs, our solver introduces a novel improvement by incorporating a tiny GNN to learn parameters. This approach not only preserves graph information but also maintains the simplicity and lightweight characteristics of neural solvers.   
    \item For large-scale graph tasks, we integrate custom-designed sparsification and storage compression methods into the solver’s architecture. These tailored techniques enable efficient processing of large-scale graphs, improving performance and reducing overhead.
    \item We validate our approach through extensive experiments on nine real-world datasets, including four large-scale. Our results demonstrate that IGNN-Solver incurs minimal computational overhead (approximately 1\% of the total) and achieves up to a $8$-fold increase in inference speed without compromising accuracy.
\end{itemize}

\section{Related Works}
\label{sec2}
The typical GNN~\cite{GCN} and its variants~\cite{GAT, graphsage} have been widely used for graph data modeling in various tasks. Different GNNs have been proposed to utilize attention mechanism~\cite{GAT}, neighbors sampling~\cite{graphsage}, pseudo-coordinates~\cite{MoNet} and graph fusion~\cite{AMGCN}. However, due to issues such as over-smoothing, depth, and bottlenecks, these models typically involve finite aggregation layers~\cite{li2018deeper, alon2020bottleneck, zhang2023comprehensive}. To address it, recent works~\cite{IGNN, EIGNN, MIGNN} have developed Implicit Graph Neural Networks, encouraging these models to capture long-range dependencies on graphs. Here, we highlight the contributions of our proposed IGNN-Solver through a detailed comparison with Implicit Graph Neural Networks (IGNNs) and Deep Equilibrium Models (DEQs), both of which are closely related to our approach.

\subsection{Implicit Graph Neural Networks}
Instead of stacking a series of operators hierarchically, implicit GNNs define their outputs as solutions to nonlinear dynamical systems, which is initially introduced by~\cite{IGNN} to tackle challenges associated with learning long-range dependencies in graphs.
\cite{liu2024scalable} proposes a new implicit graph model enabling mini-batch training without sacrificing the ability to capture long-range dependencies.
Subsequently,~\cite{liu2022mgnni} introduces a novel approach based on implicit layer to model multi-scale structures on graphs.
Additionally,~\cite{MIGNN} theoretically investigates the well-posedness of the IGNN model from a monotone operator viewpoint.

Although the aforementioned IGNN works well by alleviating the problem of over-smoothing of features by allowing meaningful fixed points to propagate implicitly, the inherent \textit{slow inference speed} of implicit networks poses a major obstacle to its scalability. The main reason is that the solver of the fixed-point network is inefficient (e.g., the Picard solver used by IGNN~\cite{IGNN} and Anderson acceleration solver used by MIGNN~\cite{MIGNN}), makes the overhead of the fixed-point solver magnified by the task scales. In comparison, our proposed IGNN-Solver accelerates the iteration process of IGNNs, which addresses \textit{the most limiting drawback} compared to traditional feedforward models.

Another class of IGNNs based on Neural ODEs~\cite{chen2018neural} has emerged to address issues like depth and bottlenecks. For example, \cite{avelar2019discrete} models continuous residual layers using GCNs. \cite{poli2019graph} proposes methods for modeling static and dynamic graphs with GCNs, along with a hybrid approach where latent states evolve continuously between RNN steps in dynamic graphs. \cite{xhonneux2020continuous} tackles the problem of continuous message passing. \cite{chamberlain2021grand} introduces a graph neural diffusion network based on the discretization of diffusion PDEs on graphs. \cite{thorpe2022grand++} enhances graph neural diffusion with a source term and connects the model to random walk formulation on graphs.
However, they essentially view deep learning on graphs as a continuous diffusion process, which differs from IGNNs based on fixed-point networks, thus requiring manually tuning the termination time and step size of the diffusion equation.
In contrast, our IGNN-Solver uses implicit formulations instead of explicit diffusion discretization, which admits an equilibrium corresponding to infinite diffusion steps and expands the receptive field.


\subsection{Fixed-point Solvers for Deep Equilibrium Models}
Traditional deep learning models use multi-layered networks, while DEQs~\cite{DEQ} find a fixed-point of a single layer, representing the network's equilibrium state. Viewed as infinitely deep networks, DEQs use root-finding for forward propagation and implicit differentiation for backward propagation. Therefore, DEQs capture complex patterns with constant memory and computational costs~\cite{ling2023global}.
Monotone operator theory has been used to guarantee the convergence of DEQs~\cite{winston2020monotone} and to improve the stability of implicit networks~\cite{bai2021neural}.

However, it is well-known that DEQs, as typical implicit models, suffer from slow training and inference speeds~\cite{ling2024deep,liu2024scalable}, which is highly disadvantageous for large-scale scenarios like GraphGPT~\cite{zhao2023graphgpt}.
To alleviate this issue, recent efforts have explored certain improved solver methods for DEQs, further optimizing root-finding problems and making these models easier to solve.
\cite{huang2021textrm} discusses the amortization of the cost of the iterative solver that would otherwise make implicit models slow.
\cite{geng2021training} proposes a novel gradient approximation method for implicit models,  which significantly accelerates the backward passes in training implicit models.
\cite{bai2021neural} discusses the superiority of learnable solvers over generic solvers in implicit models using a tiny neural network and significantly enhances the efficiency of such models through custom neural solvers.

In contrast,
IGNN has a similar network representation to DEQ, both defining the output as \textit{the solution of the equation} to obtain network outputs.
But the difference lies in the fact that IGNN's equilibrium equations encode graph structure while DEQ does not, which will undoubtedly deepen its weakness of \textit{slow inference speed} and make IGNN's solution slower.
Insight on it, our proposed IGNN-Solver leverages graph information to guide solver acceleration, achieving \textit{fast and meaningful} implicit graph network propagation, especially in the case of graph data being large-scale.


\section{Preliminaries}
\label{sec3}
\subsection{Explicit GNNs}
Let $\mathcal{G} = \{\mA, \mX\}$ represent an undirected graph, where $\mA \in \mathbb{R}^{n \times n}$ is the adjacency matrix indicating the relationships between the nodes in $\mathcal{V} = \{ \vv_1, \vv_2, \ldots, \vv_n \}$, and $n$ is the number of nodes. The node feature matrix $\mX = [ \vx_1, \vx_2, \ldots, \vx_n ] \in \mathbb{R}^{n \times d}$ contains the features of the nodes, with $d$ representing the feature dimension.
Conventional (explicit) GNNs~\cite{GCN, GAT} feature a learnable aggregation process centered on the message-passing operation within the graph~\cite{gilmer2017neural}. This process iteratively propagates information from each node to its neighboring nodes. The formal general structure for each layer $l$ is defined as follows:
\begin{equation}\label{eq_GNN}
	\mH^{[l+1]}=f_\theta ( {\mA}, \mH^{[l]}),\quad {\mH^{[0]}}=\mX,
\end{equation}
where ${\mH^{[l]}}$ represents the hidden node representation, $f_\theta$ denotes the parameters in $l$-th layer.
A commonly used GNN is Graph Convolutional Network (GCN)~\cite{GCN}, defined as $\mH^{[l+1]}=\sigma(\widehat{\mA} \mH^{[l]} \mW^{[l]})$, where ${\mW^{[l]}}$ denotes the weight matrix of the $l$-th layer, $\sigma(\cdot)$ denotes the activation function, and ${\widehat{\mA}}={\tilde{\mD}^{-1/2}}(\mA+\mI){\tilde{\mD}^{-1/2}}$ represents the symmetric normalized graph matrix. $\tilde{\mD}$ is a diagonal matrix with $\tilde{\mD}_{ii} = 1 + \sum\nolimits_j \mA_{ij}$. GNNs leverage the above message-passing operation in  ~\eqref{eq_GNN} to learn useful information. Still, they often involve a limited number of $l$ layers due to over-smoothing~\cite{alon2020bottleneck}, making it challenging for GNNs to capture the long-range dependency (LRD) on graphs.

\subsection{Implicit GNNs}
\subsubsection{Architecture}
Similar to traditional explicit GNNs, IGNNs~\cite{IGNN, EIGNN, MIGNN, chen2022efficient} also involve an aggregation process, with the distinction that the depth of layers (iteration step $k$) is \textit{infinite}. The aggregation process in IGNNs is typically defined as
\begin{equation}\label{eq_IGNN_agg_process}
\mZ^{[k+1]}=\sigma( \mW \mZ^{[k]} \widehat{\mA} + b_{\Omega}(\mX)), k=1,2, \ldots, \infty,
\end{equation}
where $b_{\Omega}$ represents affine transformation parameterized by $\boldsymbol{\Omega}$, and the weight matrices $\mW$ and $\boldsymbol{\Omega}$ are globally shared at each iteration step.
The IGNN model $f_{\theta}$ is formally described by 
\begin{equation}\label{eq_IGNN_model}
\mZ^{\star} = f_{{\theta}}(\mZ^{\star}, \widehat{\mA}, \mX ) =\sigma(\mW \mZ^{\star} \widehat{\mA}+ b_{\Omega}(\mX)),
\end{equation}
where the representation, given as the ``internal state'' $\mZ^{\star}$, is obtained as the fixed-point solution of the equilibrium~\eqref{eq_IGNN_agg_process}.
The final representation theoretically encompasses information from all neighbors in the graph. Therefore, IGNNs capture LRD within the graph, offering better performance compared to GNNs with finite iterations~\cite{IGNN, EIGNN}. Another notable advantage of this framework is its memory efficiency, as it only needs to retain the current state $\mZ$ without requiring additional intermediate representations.

\subsubsection{Well-posedness}
Notably, to obtain a valid representation for any given input $\mX$ from IGNNs, we need to solve the unique internal state $\mZ^{\star}$ in~\eqref{eq_IGNN_model}. However, there may not always exist a well-defined solution 
 $\mZ^{\star}$ for some  inputs $\mX$~\cite{IGNN}.
To guarantee the existence and uniqueness of the solution to~\eqref{eq_IGNN_model}, we introduce the concept of \textit{well-posedness} for IGNNs $f_{\theta}$ with activation function $\sigma$~\cite{IGNN}. Specifically, the weight matrix $\mW \in$ $\mathbb{R}^{m \times m}$ and the adjacency matrix $\widehat{\mA} \in \mathbb{R}^{n \times n}$  are said to be well-posed for $\sigma$  if  the solution $\mZ^{\star} \in \mathbb{R}^{m \times n}$ of ~\eqref{eq_IGNN_model} exists and is unique for any $b_{\Omega}(\mX) \in \mathbb{R}^{m \times n}$.



Based on the \textit{monotone operator theorem}~\cite{ryu2016primer, winston2020monotone},  a sufficient condition for the well-posedness for IGNNs is defined as follows.
Let the nonlinearity $\sigma$ be a non-expansive activation function (\textit{e.g.}, Sigmoid, Tanh, ReLU, \textit{etc}.), and let $\widehat{\mW}=\widehat{\mA}^{\top} \otimes \mW$. 
\footnote{We represent the Kronecker product of matrices $\mA$ and $\mB$ as $\mA \otimes \mB$. }
Then, the IGNN model defined in~\eqref{eq_IGNN_model} is well-posed as long as $\mI-\widehat{\mW} \succeq m \mI $ for some $m>0$.
\footnote{ We represent $\mA \succeq \mB$ if $\mA-\mB$ is semi-positive definite. For the case of non-symmetric matrices $\widehat{\mW}$, positive definiteness can be defined as the positive definiteness of the symmetric component, i.e., $\mI-\widehat{\mW} \succeq m \mI$ if and only if $\mI-\frac{\widehat{\mW}+\widehat{\mW}^T}{2} \succeq m \mI$.}


It establishes that if $\widehat{\mW}$ is symmetric and $\mI-\widehat{\mW} \succeq m \mI$ for some $m>0$, then the existence and uniqueness of the equilibrium point $\mZ^{\star}$ are guaranteed, ensuring the model is well-posed. This follows from the sufficient condition that $\mI - \widehat{\mW}$ is strongly monotone. 

To satisfy the well-posedness of IGNNs, a commonly used method~\cite{jing2017tunable, MIGNN}, is to   orthogonally reparameterize $\mW$ in the IGNN layer $f_{\theta}$ using the following Cayley map, which is inspired from the unitary operator~\cite{halmos1982hilbert},  as
\begin{equation}\label{eq_Cayley_map}
    \mW=\kappa(\mI-\mS)(\mI+\mS)^{-1},
\end{equation}
where $\kappa \in(0,1)$ and $\mS=\mC-\mC^{\top}$ is a skew-symmetric matrix with $\mC$ as an arbitrary parameterized matrix.
From~\eqref{eq_Cayley_map}, it follows that $\mW$ is positive definite and its real eigenvalues are in $(0,1)$. Moreover, based on the definition of $\widehat{\mA}$, we can derive the eigenvalues of $\widehat{\mW}$ are in $(0,1)$, which ensures that $\mI-\widehat{\mW} \succeq m \mI $, thereby ensuring the well-posedness of IGNN. We refer interested readers to~\cite{MIGNN} for detailed proof.



\subsection{Training of IGNNs}
Additionally, to train the parameters $\boldsymbol{\theta} = \{\mW, \boldsymbol{\Omega}\}$ during the back-propagation process of neural networks, IGNNs compute the Jacobian $\nabla_\mZ \mathcal{L}$ by solving the equilibrium equation, as derived from the implicit differential method and the \textit{implicit function theorem}~\cite{krantz2002implicit}:
\begin{equation}\label{eq_Jacobian}
    \nabla_\mZ \mathcal{L}=\mD \odot( \mW^{\top} \nabla_\mZ \mathcal{L} \widehat{\mA}^{\top}+\nabla_\mX \mathcal{L}),
\end{equation}
where $\mD=\sigma^{\prime}(\mW \mX \widehat{\mA}+b_{\Omega}(\mU))$ and $\sigma^{\prime}(\cdot)$ refers to the element-wise derivative of the map $\sigma$. Once $\nabla_\mZ \mathcal{L}$ is obtained, $\nabla_\mW \mathcal{L}$ and $\nabla_{\boldsymbol{\Omega}} \mathcal{L}$ are derived via the chain rule, eliminating the need to store activations at each layer during gradient updates.

However, IGNNs typically require multiple iterations to reach equilibrium, both in finding the fixed point during the forward process \eqref{eq_IGNN_agg_process} and in solving the implicit differential during the backward process \eqref{eq_Jacobian}, resulting in significant overhead during both training and inference. Therefore, developing \textit{a fast and efficient solver} to find their fixed points is crucial.

\subsection{Traditional Anderson Acceleration Solver}

\begin{algorithm*}
	\renewcommand{\algorithmicrequire}{\textbf{Input:}}
	\renewcommand{\algorithmicensure}{\textbf{Output:}}
	\caption{Anderson Acceleration Solver}
	\label{AA}
	\begin{algorithmic}[1]
		\Require fixed-point function $f_\theta: \mathbb{R}^n \rightarrow \mathbb{R}^n$, max storage size $m$, residual control parameter $\beta$

		\State Set initial point value $\mZ^{[0]} \in \mathbb{R}^n$
		\Comment{set $\mathbf{0}$ or random value normally}

		\For{$k=0$, $\ldots$, $K-1$}

		\State 1) Set $m_k=\min \{m, k\}$
		\Comment{storage of the most recent steps}

		\State 2) Solve $\alpha^{[k]}=\arg \min _{\alpha \in \mathbb{R}^{m_k+1}}\left\|\mG^{[k]} \alpha\right\|_2 \text {, s.t. } \mathbf{1}^{\top} \alpha^{[k]}=1$
		\Comment{compute weights}

		\State 3) $\mZ^{[k+1]}=\beta \sum_{i=0}^{m_k} \alpha_i^{[k]} f_\theta(\mZ^{[k-m_k+i]})+(1-\beta) \sum_{i=0}^{m_k} \alpha_i^{[k]} \mZ^{[k-m_k+i]} \quad$
		\Comment{Anderson step}

		\EndFor


	\end{algorithmic}
\end{algorithm*}

Since implicit networks typically require multiple iterations to reach equilibrium, a commonly used solver method is Anderson Acceleration~\cite{anderson1965iterative}, which is widely applied in~\cite{MIGNN, MDEQ, DEQ, bai2021neural}. Specifically, Given an implicit network layer $f_\theta$ and input $\mX$, we aim to find the fixed-point $\mZ^*$ in the system that represents the equilibrium state of the network output by solving a root-finding problem
\begin{equation}\label{eq_DEQ}
	g_\theta\left(\mZ^{\star}, \mX\right):=f_\theta\left(\mZ^{\star}, \mX\right)-\mZ^{\star}=0,
\end{equation}
via utilizing fixed-point solvers, such as the classical Anderson Acceleration (AA)~\cite{anderson1965iterative}, which can directly seek the equilibrium point $\mZ^{\star}$ through quasi-Newton methods. This approach demonstrates super-linear convergence properties.

We briefly introduce the AA-solver, as our approach is inspired by it. Algorithm~\ref{AA} provides a pseudocode example, where $\mG^{[k]}=[g_\theta(\mathbf{z}^{[k-m_k]}), \ldots, g_\theta(\mathbf{z}^{[k]})]$ represents the past residuals. The key idea behind AA is to accelerate the next approximate solution by leveraging information stored from the history $m$ iteration steps, constructing a normalized linear combination of past updates with weights $\alpha^{[k]}$ and $\beta$. These weights are computed greedily to minimize the linear combination of past AA update steps. This approach is typically effective for accelerating the solution of implicit functions.


However, since the computation process relies only on the final output and does not require storing any intermediate activation values~\cite{DEQ}, the memory consumption during the training process remains constant, which is also a key reason why fixed-point solvers are attractive.

\begin{figure*}
\centering
\includegraphics[width=0.75\linewidth]{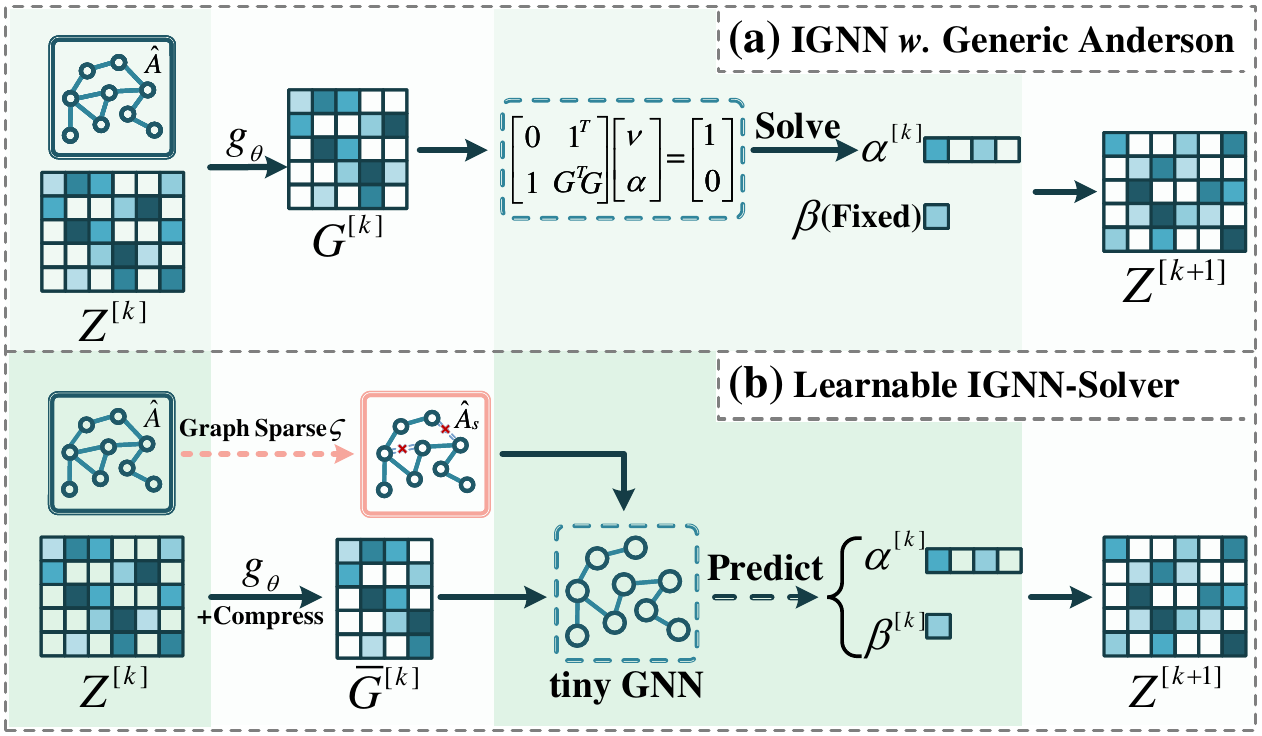}
\caption{
    Comparison of our IGNN-Solver and the generic Anderson solver at each iteration. 
    (a) The canonical Anderson solver performs local least-squares fitting at each iteration, with a fixed $\beta = \beta^{[k]}$ set to a constant.
    (b) Our learnable IGNN-Solver improves and incorporates learnable iterative updates for enhanced convergence via tiny GNN.
}
\label{FlowChart}
\end{figure*}

\section{Our purposed IGNN-Solver}
\label{sec4}
Although traditional fixed-point solvers for IGNNs demonstrate functionality, they are characterized by relatively slow performance, necessitate manual parameter tuning, and are not well-suited for graph-based applications. 
To address these limitations, we propose a lightweight, learnable, and content-aware fixed-point solver for IGNN that incorporates a tiny graph neural network, combining the speed advantages of the solver with the information benefits of graph learning. By encoding the relationship between past residuals and parameters $\alpha^{[k]}$ and $\beta$, it enables more efficient adjustments to enhance the learning capabilities of the IGNN-Solver, as shown in Figure~\ref{FlowChart}.

Given the implicit model's characteristic that its representation capability is decoupled from the forward computation (i.e., the solver has no knowledge of the task, such as node classification, and vice versa), we can train this neural solver in a lightweight, unsupervised manner. Once learned, the IGNN-Solver can solve the fixed-point equation within the IGNN layer, facilitating further model updates. Note that our solver does not alter the original structure of the model but only accelerates the solution process. Details of the training strategy involving the IGNN-Solver are provided in Appendix~\ref{Appendix_Training_Strategies}.

\begin{algorithm*}[t]
    \renewcommand{\algorithmicrequire}{\textbf{Input:}}
    \renewcommand{\algorithmicensure}{\textbf{Output:}}
    \caption{IGNN-Solver Iterations and Training}
    \label{Algorithm_2}
    \begin{algorithmic}[1]
        \Require frozen IGNN model $f_\theta$, initializer $h_\phi$, tiny-GNN predictor $s_{\xi}$, sparse subgraph $\widehat{\mA}_s$, storage $\mG \in \mathbb{R}^{(m+1) \times d'}$.
        \State Compute the initial value of fixed point by $\mZ^{[0]}=h_\phi(\mX) \in \mathbb{R}^{d'}$
        \Comment{initializer to guess initial value}
    
        \State Define $g_\theta(\mZ)=f_\theta(\mZ)-\mZ$ and set $\mG[0]=g_\theta(\mZ^{[0]})$
    
        \For{$k=0, \ldots, K$}
    
        \State 1) Set $m_k=\min \{m, k\}$ and $\mG^{[k]}=\mG \left[0:\left(m_k+1\right)\right] \in \mathbb{R}^{\left(m_k+1\right) \times d'}$
        \Comment{compress storage}
    
        \State 2) Predict ${\alpha}^{[k]}, {\beta}^{[k]} = s_{\xi} ( \mG^{[k]}, \widehat{\mA}_s )$, where ${\alpha}^{[k]} \in \mathbb{R}^{(m_k+1)}$, $\mathbf{1}^{\top} {\alpha}^{[k]} = 1$ and $\beta^{[k]} \in \mathbb{R}^{1}$
        \Comment{predict with tiny-GNN}

    
        \State 3) Update $\mZ^{[k+1]}=\beta^{[k]} \cdot \mathbf{1}^{\top} \mG^{[k]}+(1-\beta^{[k]})\sum_{i=0}^{m_k} \alpha_i^{[k]} \mZ^{[k-m_k+i]}$
        \Comment{Anderson step}
    
        \State 4) Update $\mG = \operatorname{concat}\left(\mG[1:],[g_\theta(\mZ^{[k+1]})]\right)$
    
    
        \EndFor
    
    
        \If{it is inference stage}
        \State \Return $\mZ^{[k+1]}$
    
        \Else 
        \State \Return $( \mZ^{[k+1]}, \mG^{[k]}, {\alpha}^{[k]}, {\beta}^{[k]} )_{k=0, \ldots, K}$ and $\mZ^{[0]}$
    
        \State Compute $\mathcal{L}_{\text {total }}$ and back-propagate it to update IGNN-Solver $\left\lbrace s_{\xi}, h_\phi\right\rbrace $
        \EndIf
    
    \end{algorithmic}
\end{algorithm*}

\subsection{General Formulation}
Let $\mX \in \mathbb{R}^{d \times n}$ be the input features and $\widehat{\mA} \in \mathbb{R}^{n \times n}$ be the graph, IGNN learns the node representation by finding the fixed point below
\begin{equation}\label{IGNN_formulation}
    \mZ = f_{\theta}(\mZ, \widehat{\mA}, \mX) = \sigma(\mW \mZ \widehat{\mA} +b_{\Omega}(\mX))
\end{equation}

The overall structure of the IGNN-Solver is shown in Algorithm~\ref{Algorithm_2}. Specifically, it learns the parameters $\alpha$ that control the weights among past $m$ steps approximate solution, and $\beta$ that control the weights between past residuals and approximate solution, through a solver $s_{\xi}$ to accelerate the approximate solution of the next update step.
To train the IGNN-Solver, we minimize the joint loss function $\mathcal{L}_\text{total}$ by back-propagating through this $K$-step temporal process, which is discussed in Section~\ref{sec4.2}.
It's worth noting that the original IGNN is frozen (i.e., model parameters $\theta$ are fixed), and only the IGNN-Solver parameters $\xi$ are trained here, so we do not need the ground-truth label $y$ that corresponds to input $x$. This implies that IGNN-Solver can also be fine-tuned during inference after deployment.


\subsubsection{Initializer}\label{sec4.1.1}
To accelerate the convergence of predicted values, we propose constructing an initializer 
\begin{equation}
    \mZ^{[0]}=h_\phi(\mX) 
\end{equation}
 where $h_\phi: \mathbb{R}^d \rightarrow \mathbb{R}^{d'}$, for a rapid and reasonable input-based initial estimate value, rather than simply setting the initial values to $\mathbf{0}$ or random, where $\phi$ are learnable parameters of the initializer.
We set the intermediate dimension $d'$ to be significantly smaller than the feature dimension $d$ to reduce the training overhead of the initializer.




\subsubsection{Improved Anderson Iterations with tiny-GNN}
In the original AA solver, the parameter $\beta$ is predetermined and fixed~\cite{anderson1965iterative}, whereas the parameters  $\alpha^{[k]}$s are determined by solving numerous linear equations using the least squares method. Although this method operates adequately, its efficiency and adaptability in optimizing IGNN models are limited.
Therefore, we propose introducing a tiny and learnable graph neural network as
\begin{equation}\label{4}
	\alpha, \beta = s_{\xi}(\mG, \widehat{\mA}) \,
\end{equation}
where $\, s_{\xi} :(\mathbb{R}^{(m_k+1) \times d'} \times \mathbb{R}^ {n} ) \rightarrow (\mathbb{R}^{(m_k+1)} \times \mathbb{R}^1)$, to predict the two parameters instead of setting them as the least squares solution on the past residuals $\mG$.

\begin{remark}[Well-posedness of IGNN-Solver]\label{remark_Well-posedness_of_IGNN_Solver}
The proposed IGNN-Solver remains well-posed when the IGNN satisfies the reparameterization defined in~\eqref{eq_Cayley_map}.
Note that the IGNN-Solver introduces two key improvements while maintaining the structure and parameters of the model. First, we propose an initializer for a better initial guess, establishing a more meaningful starting point, reducing solving time, and shortening the fixed-point trajectory. Second, we parameterize the Anderson weights and step sizes within their original range, ensuring that well-posedness is preserved. Consequently, the modifications introduced in the IGNN-Solver retain the foundational characteristics of classical AA \textit{without compromising the well-posedness condition}. Therefore, for an IGNN $(f_{\theta}, b_{\Omega}, \sigma )$ where $f_{\theta}$ satisfies the well-posedness condition, it holds that the corresponding IGNN-Solver $(h_\phi, s_{\xi})$ on this IGNN is also well-posed, thereby guaranteeing the existence and uniqueness of its solution.
\end{remark}



\subsection{Graph Sparsification and Storage Compression}
Directly using the original graph for modeling would \textit{result in an oversized network}~\cite{RPI-Graph, zheng2020robust}, deviating from the initial goal of using a tiny network for prediction. Moreover, because of \textit{the curse of dimensionality}, directly mapping the data from high to extremely low dimensions is not appropriate~\cite{koppen2000curse}.
Therefore, in response to the challenges posed by large-scale graphs, we take into account both Graph Sparsification and Storage Compression in the design of the solver, as detailed below.
\subsubsection{Graph Sparsification}
Firstly, the number of edges in graph $\widehat{\mA}$ is usually large in practice, as it is affected by the scale of the input (for example, in the ogbn-products dataset, the number of edges in the graph is close to 62M).
To maintain the lightweight nature of tiny-GNN and reduce the computational cost of prediction, we propose introducing graph sparsification~\cite{hui2023rethinking, lutzeyer2022sparsifying, zhang2023joint} for a  light graph:
\begin{equation}
	\widehat{\mA}_s = \varsigma( \widehat{\mA}, \beta ),
\end{equation}
where $\widehat{\mA}_s$ represents the sparse subgraph of $\widehat{\mA}$, obtained via a graph sparsification operator $\varsigma$, with the parameter $\beta$ controls the sparsity of $\widehat{\mA}_s $. We choose the RPI-Graph approach~\cite{RPI-Graph} in this paper, which is a plug-and-play graph sparsification method based on the principle of relevant information.
It is worth noting that this subgraph is used exclusively in the tiny-GNN predictor $s_{\xi}$ and does not alter the original graph of the IGNN model $f_\theta$.

\subsubsection{Storage Compression}
Besides, considering that $n$ (the number of nodes) is relatively large (for example, $n$ is approximately 169K in the ogbn-arxiv dataset), it is not appropriate to map $s_{\xi}$ from a high-dimensional space to a very low-dimensional one in~\eqref{4} directly~\cite{AMGCN, zhao2021heterogeneous, poli2020hypersolvers} owing to the curse of dimensionality~\cite{koppen2000curse}.
Therefore, to maintain $s_{\xi}$ fast and compact, we recommend compress $\mG^{[k]}_i$ to form a smaller yet still representative version $\bar{\mG}^{[k]}_i$. Specifically, We map it from $\mathbb{R}^{(m_k+1) \times d'}$ to an appropriate space $\mathbb{R}^{(m_k+1) \times p}$ by multiple layer perception (MLP). Then, the nearest $m_k+1$ sets of fixed points are merged into one feature matrix:
\begin{equation}
	\bar{\mG}^{[k]} = \mathop{||}_{i=k-m_k}^k  \bar{\mG}^{[k]}_i,
\end{equation}
where $ \mathop{||}_{i=k-m_k}^k $ represents the concatenation operation that stacks the compressed storage matrix $\bar{\mG}^{[k]}_i \in \mathbb{R}^{(m_k+1) \times p}$ in the $k$-th iteration.

Once the smaller yet representative storage matrix $\bar{\mG}$ is obtained as described above, we treat it as a $p$-channel matrix and employ a tiny graph neural network layer:
\begin{equation}
	\alpha^{[k]}, \beta^{[k]} = s_{\xi} ( \bar{\mG}^{[k]}, \widehat{\mA}_s ),
\end{equation}
to predict the relative weights $\alpha^{[k]}$ assigned to past $m_k+1$ residual, as well as the AA mixing coefficient $\beta^{[k]}$ at the $k$-th iteration. Notably, we normalize the original output $\widehat{\alpha}^{[k]}$ using the function 
\begin{equation*}
    \alpha^{[k]}=\widehat{\alpha}^{[k]}+\frac{\left(1-\mathbf{1}^{\top} \widehat{\alpha}^{[k]}\right)}{m_k+1} \cdot \mathbf{1}
\end{equation*} 
to ensure that the constraint  $\mathbf{1}^{\top} \alpha^{[k]}=1$ is satisfied.

In this way, $s_{\xi}$ shall autonomously learn and adjust these parameters $\alpha^{[k]}$ and $\beta^{[k]}$ based on previous solver steps, while also receiving gradients from subsequent iterations. We provide a comparison of different choices for $s_{\xi}$ in Section~\ref{sec_others}, demonstrating that IGNN-Solver improves the convergence path. We also introduce the training procedure below.

\subsection{Training IGNN-Solver}\label{sec4.2}
Unlike the neural ODE solver, the fixed-point trajectory of the IGNN is not unique. Therefore, trajectory fitting is not applicable to the IGNN-solver training. Instead, the goal is simply to bring everything as close as possible to $\mZ^\star$. Formally, given an IGNN-Solver $\left\lbrace s_{\xi}, h_\phi\right\rbrace $ that returns $( \mZ^{[k+1]}, \mG^{[k]}, {\alpha}^{[k]}, {\beta}^{[k]} )_{k=0, \ldots, K}$ and $\mZ^{[0]}$, we introduce three objectives functions for its training. The complete training algorithm can be referred to in Algorithm~\ref{Algorithm_2}.


\paragraph{Initializer Loss Functions}
To train the initializer, we minimize the distance between the initial guess and the fixed-point by
\begin{equation}
    \mathcal{L}_{\text {init }}=\left\|h_\phi(\mX)-\mZ^{\star}\right\|_2,
\end{equation}
for facilitating the subsequent iteration process. Since the initializer directly predicts based on the input $\mX$ without going through iteration, we separate this loss from other components.

\paragraph{Reconstruction Loss Functions}
The training of the solver does not require reference to label information or any trajectory fitting. Apart from the loss $\mathcal{L}_{\text{init}}$ necessary for training the initializer above, we introduce reconstruction loss
\begin{equation}
	\mathcal{L}_{\text {rec}}^{[k]} = \|\mZ^{[k]}-\mZ^{\star}\|_2, \quad
\end{equation}
where the reconstruction loss $\mathcal{L}_{\text {rec }}^{[k]}$ aims to make all intermediate predictions $\mZ^{[k]}$ converge to the accurate fixed point $\mZ^{\star}$ as closely as possible.

\paragraph{Auxiliary Loss Functions}
Although we utilize $\alpha^{[k]}$ and $\beta^{[k]}$ to improve the generic solver, we have empirically found that using an auxiliary loss to guide the solver's prediction of $\alpha$ is beneficial sometimes, especially in the early stages of training. Therefore, in practice, we suggest considering this loss and gradually diminishing it as training progresses (i.e., decay the weight of this loss to $0$).
\begin{equation}
	\mathcal{L}_{\alpha} = \sum_{k=0}^K\|\mG^{[k]} \alpha^{[k]}\|_2,
\end{equation}

The final form of the joint loss function is as follows:
\begin{equation}\label{eq10}
	\mathcal{L}_{\text {total }} =  \lambda_1 \mathcal{L}_{\text {rec }}^{[k]} + \lambda_2 \mathcal{L}_{\text {init }} +  \lambda_3 \mathcal{L}_{\alpha },
\end{equation}
where $ \lambda_1$, $\lambda_2$ and $\lambda_3$ control the weights of three loss functions mentioned above and elaborate more in Section~\ref{sec_Hp_setting}.


\begin{figure*}
    \centering
    \begin{subfigure}{4cm}
        \includegraphics[width=1\textwidth]{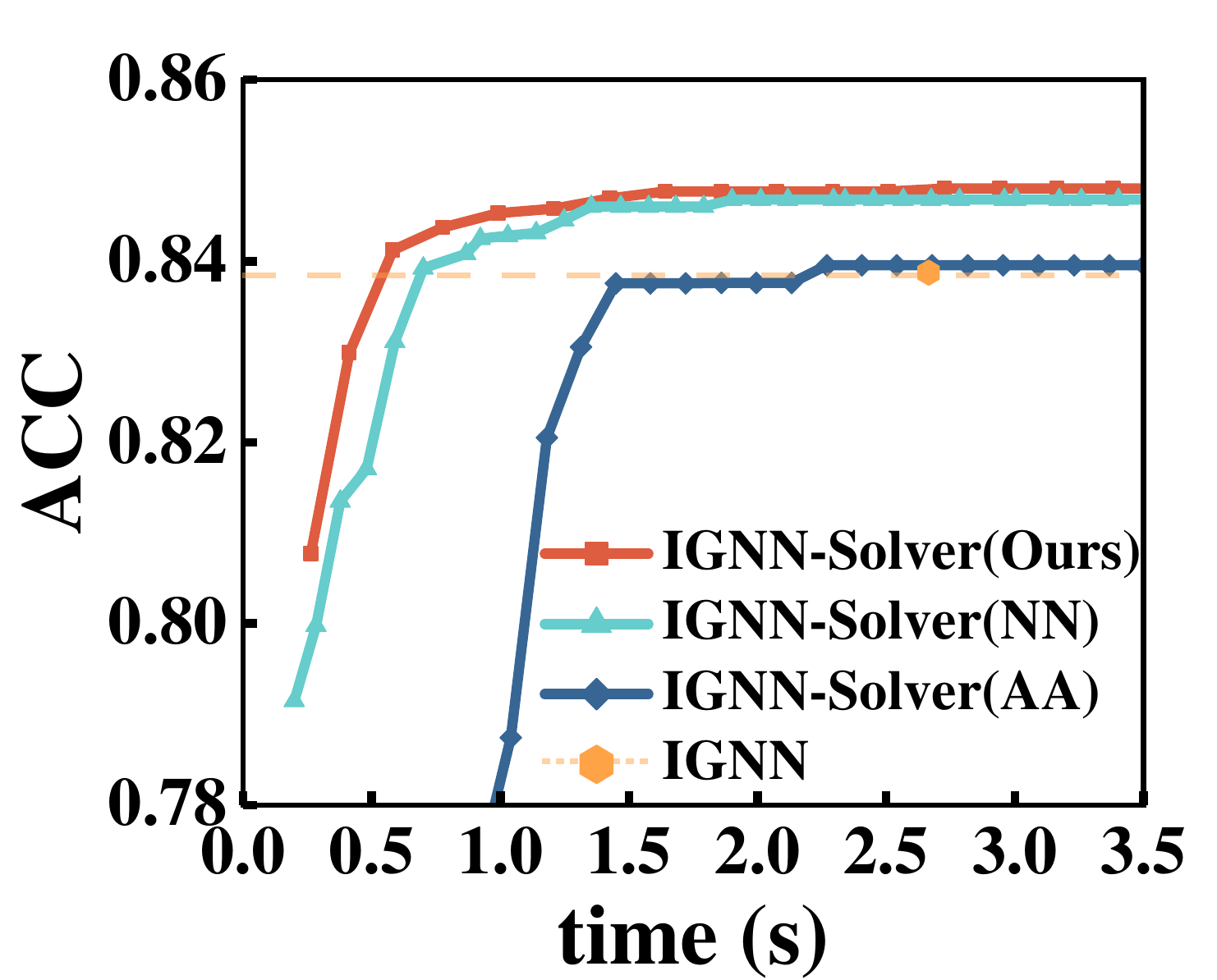}
        \caption{Amazon-all}
        \label{t_Amazon-all}
    \end{subfigure}
    \begin{subfigure}{4cm}
        \includegraphics[width=1\textwidth]{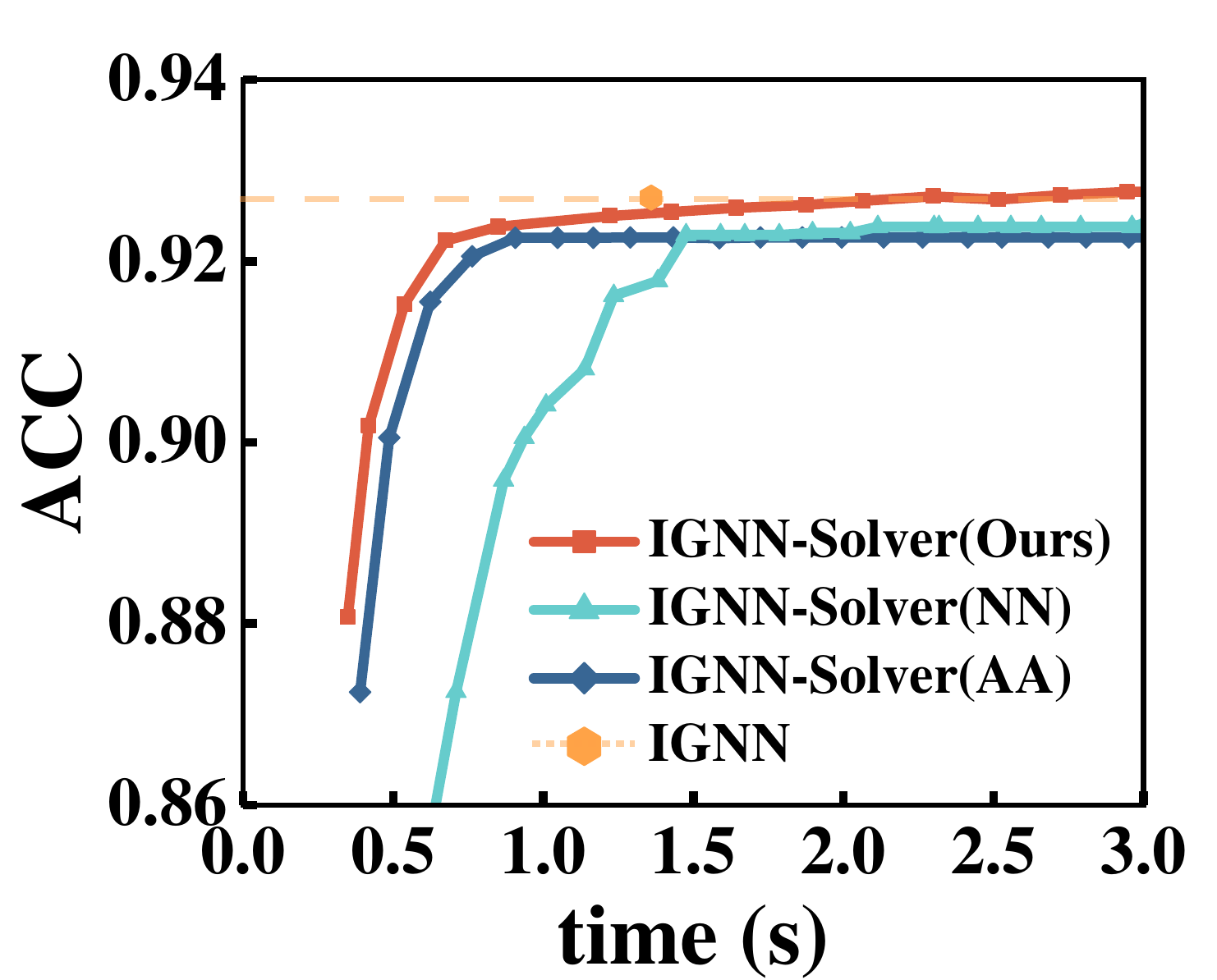} 
        \caption{Reddit}
        \label{t_Reddit}
    \end{subfigure}
    \begin{subfigure}{4cm}
        \includegraphics[width=1\textwidth]{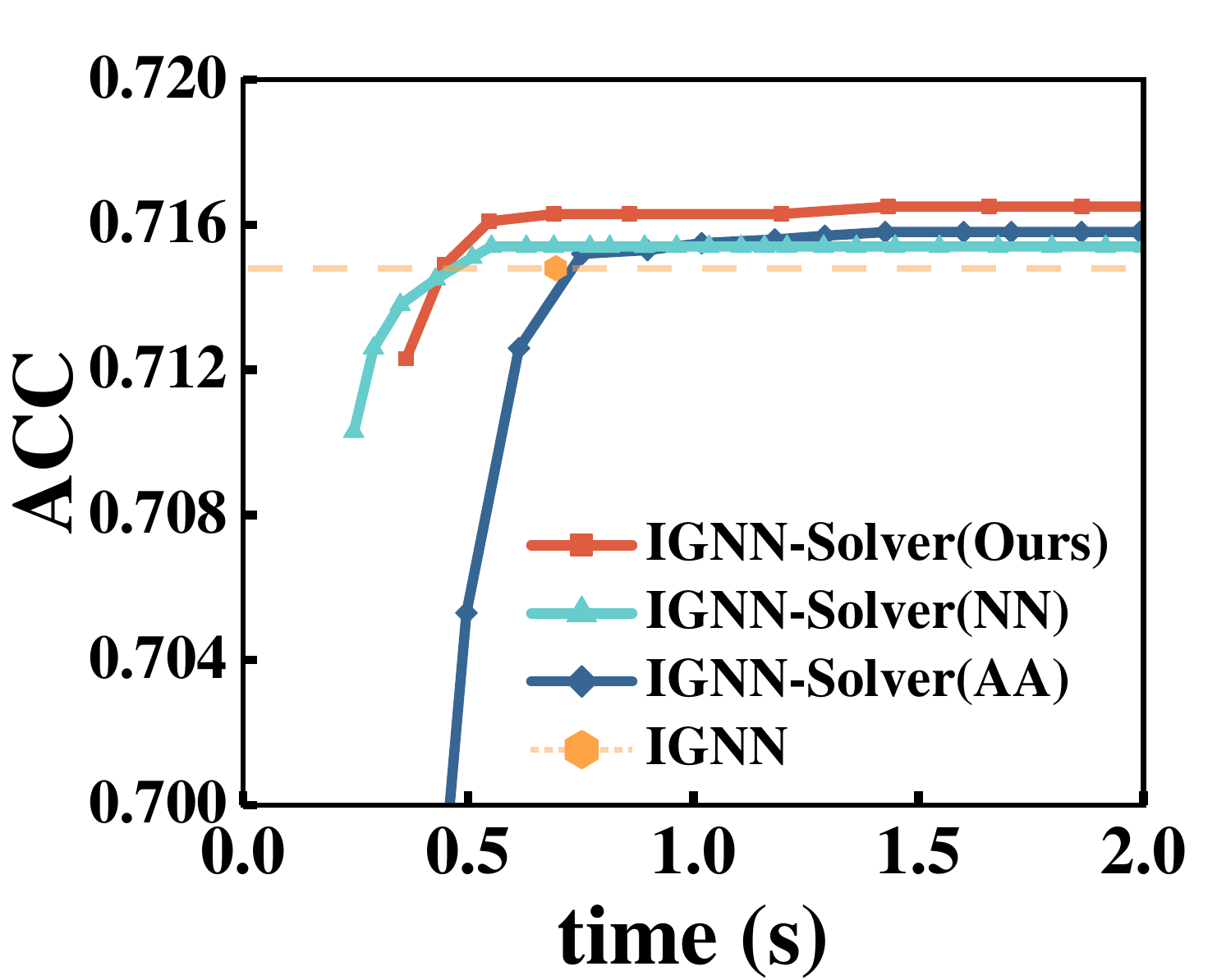} 
        \caption{ogbn-arxiv}
        \label{t_ogbn-arxiv}
    \end{subfigure}
    \begin{subfigure}{4cm}  
        \includegraphics[width=1\textwidth]{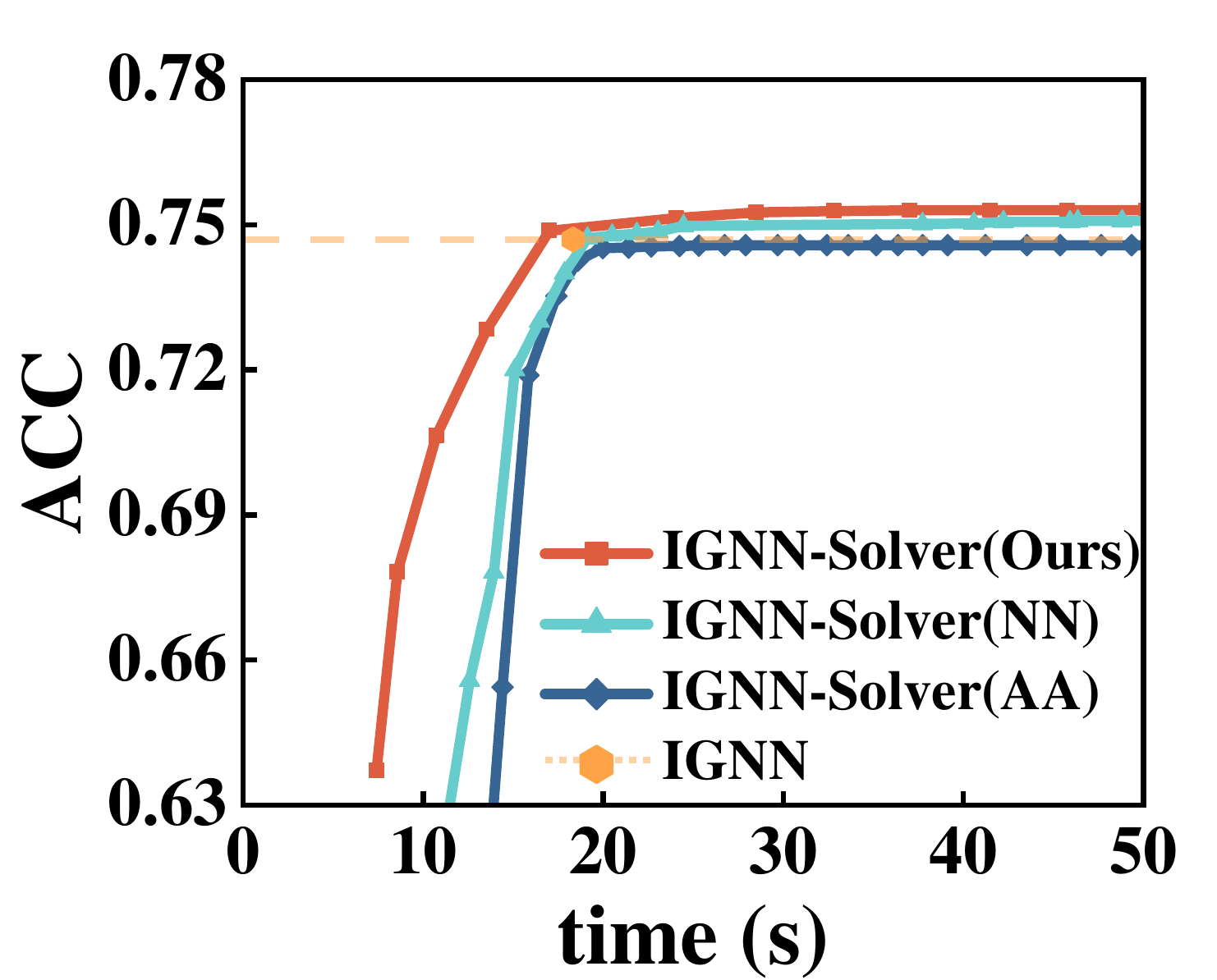} 
        \caption{ogbn-products}
        \label{t_ogbn-products}
    \end{subfigure}
    \caption{Speed-accuracy tradeoff curves comparing the inference time of \textcolor[RGB]{255, 162, 71}{\textbf{IGNN}} with \textcolor[RGB]{222, 92, 64}{\textbf{IGNN-Solver}} and its variants \textcolor[RGB]{102, 204, 204}{\textbf{IGNN-Solver (NN)}} and \textcolor[RGB]{56, 102, 148}{\textbf{IGNN-Solver (AA)}}, across four \textit{large-scale} datasets: Amazon-all~(\ref{t_Amazon-all}), Reddit~(\ref{t_Reddit}), ogbn-arxiv~(\ref{t_ogbn-arxiv}) and ogbn-products~(\ref{t_ogbn-products}). Each plot shows the average performance over five independent runs under identical experimental conditions.}
    \label{F2}
\end{figure*}

\begin{table*}
\caption{Node classification accuracy results on four \textit{large-scale} real-world datasets (more results on small-scale datasets including Citeseer, ACM, CoraFull, BlogCatalog, Flickr can be found in Table~\ref{t1_2}), with experiments conducted over five trials. The mean accuracy (\%) ± standard deviation is reported. The \textbf{best} and the \underline{runner-up} results are highlighted in boldface and underline, respectively.}
\label{t1_1}
\centering
\begin{tabular}{clcccc}
\toprule
\multirow{3}{*}{Type}     & *Models                  & Amazon-all                   & Reddit                       & ogbn-arxiv                   & ogbn-products                \\
                          & *Nodes                   & $334,863$                    & $232,965$                    & $169,343$                    & $2,449,029$                  \\
                          & *Edges                   & $14,202,057$                 & $11,606,919$                 & $1,166,243$                  & $61,859,140$                 \\ \midrule
\multirow{9}{*}{Explicit} & GCN \cite{GCN}           & $79.12 \pm 1.11$             & $89.65 \pm 0.14$             & $71.56 \pm 0.25$             & $71.91 \pm 0.27$             \\
                          & GAT \cite{GAT}           & $76.02 \pm 2.09$             & $90.08 \pm 0.44$             & $71.10 \pm 0.24$             & $72.33 \pm 0.32$             \\
                          & SGC \cite{SGC}           & $75.56 \pm 1.75$             & $91.44 \pm 0.41$             & $64.66 \pm 1.02$             & $70.48 \pm 0.19$             \\
                          & APPNP \cite{APPNP}       & $79.80 \pm 1.22$             & $91.68 \pm 0.24$             & $71.28 \pm 0.29$             & $74.46 \pm 0.64$             \\
                          & JKNet \cite{JKNet}       & $81.19 \pm 1.09$             & $91.71 \pm 0.31$             & $71.08 \pm 0.35$             & $73.70 \pm 0.58$             \\
                          & AM-GCN \cite{AMGCN}      & $81.85 \pm 1.46$             & $90.20 \pm 0.34$             & $68.83 \pm 0.67$             & $73.19 \pm 0.49$             \\
                          & DEMO-Net  \cite{demonet} & $84.08 \pm 1.29$             & $89.53 \pm 0.29$             & $68.40 \pm 0.24$             & $73.88 \pm 0.62$             \\
                          & GCNII   \cite{GCNII}     & $83.60 \pm 2.40$             & $89.87 \pm 0.38$             & $67.66 \pm 0.20$             & $72.88 \pm 0.18$             \\
                          & ACM-GCN  \cite{ACM-GCN}  & $83.04 \pm 2.61$             & $90.37 \pm 0.42$             & $68.32 \pm 0.18$             & $71.68 \pm 0.41$             \\ \midrule
\multirow{6}{*}{Implicit} & IGNN \cite{IGNN}         & $83.90 \pm 0.51$             & $92.30 \pm 1.55$             & $70.49 \pm 0.75$             & $74.63 \pm 0.24$             \\
                          & EIGNN \cite{EIGNN}       & $\underline{84.32} \pm 0.57$ & $92.00 \pm 0.24$             & $70.59 \pm 0.31$             & $74.58 \pm 0.26$             \\
                          & MIGNN \cite{MIGNN}       & $83.68 \pm 0.82$             & $91.98 \pm 0.42$             & $\underline{71.95} \pm 0.44$ & $74.62 \pm 0.32$             \\
                          & IGNN-Solver (AA)      & $83.44 \pm 0.19$             & $92.37 \pm 0.35$             & $71.73 \pm 0.41$             & $74.61 \pm 0.28$             \\
                          & IGNN-Solver (NN)      & $84.13 \pm 1.04$             & $\underline{92.42} \pm 0.41$ & $70.78 \pm 0.37$             & $\underline{74.69} \pm 0.31$ \\
                          & \textbf{IGNN-Solver (Ours)}    & $\textbf{84.50} \pm 0.70$    & $\textbf{93.91} \pm 0.31$    & $\textbf{72.53} \pm 0.41$    & $\textbf{74.90} \pm 0.20$    \\ \bottomrule
\end{tabular}

\end{table*}

\section{Experiments}\label{sec5}
In this section, we compare the speed/accuracy Pareto curve rather than a single point on the curve of IGNN-Solver with IGNNs and several state-of-the-art (SOTA) GNNs across various graph classification tasks, both at the node and graph levels. Our goal is to demonstrate: 1) IGNN-Solver achieves nearly a $1.5\times$ to $8\times$ inference acceleration without sacrificing accuracy; and 2) IGNN-Solver is extremely compact, adding little overhead to training.
Specifically, we compare our approach with numerous representative baselines and two variants of our method: IGNN \textit{w.} AA (using the original Anderson Acceleration to speed up IGNN) and IGNN \textit{w.} NN (replacing our proposed graph neural solver with a standard neural network solver for parameter learning).

In Section~\ref{sec_others}, we sequentially demonstrate that the training overhead on IGNN-Solver is little relative to the total time overhead on IGNN. Additionally, we provide further evidence regarding the convergence and generalizability of the IGNN-Solver. We compare the training/inference time costs per epoch between IGNN and IGNN-Solver, visualizing the advantages of our approach. Furthermore, we conduct ablation studies to validate the stability of the IGNN solver and assess the effectiveness of its components. Additional experimental setups and descriptions are detailed in Appendix~\ref{secExperimental_Details}.

\begin{figure*}
    \centering
        \begin{subfigure}{3.4cm}
            \includegraphics[width=1\textwidth]{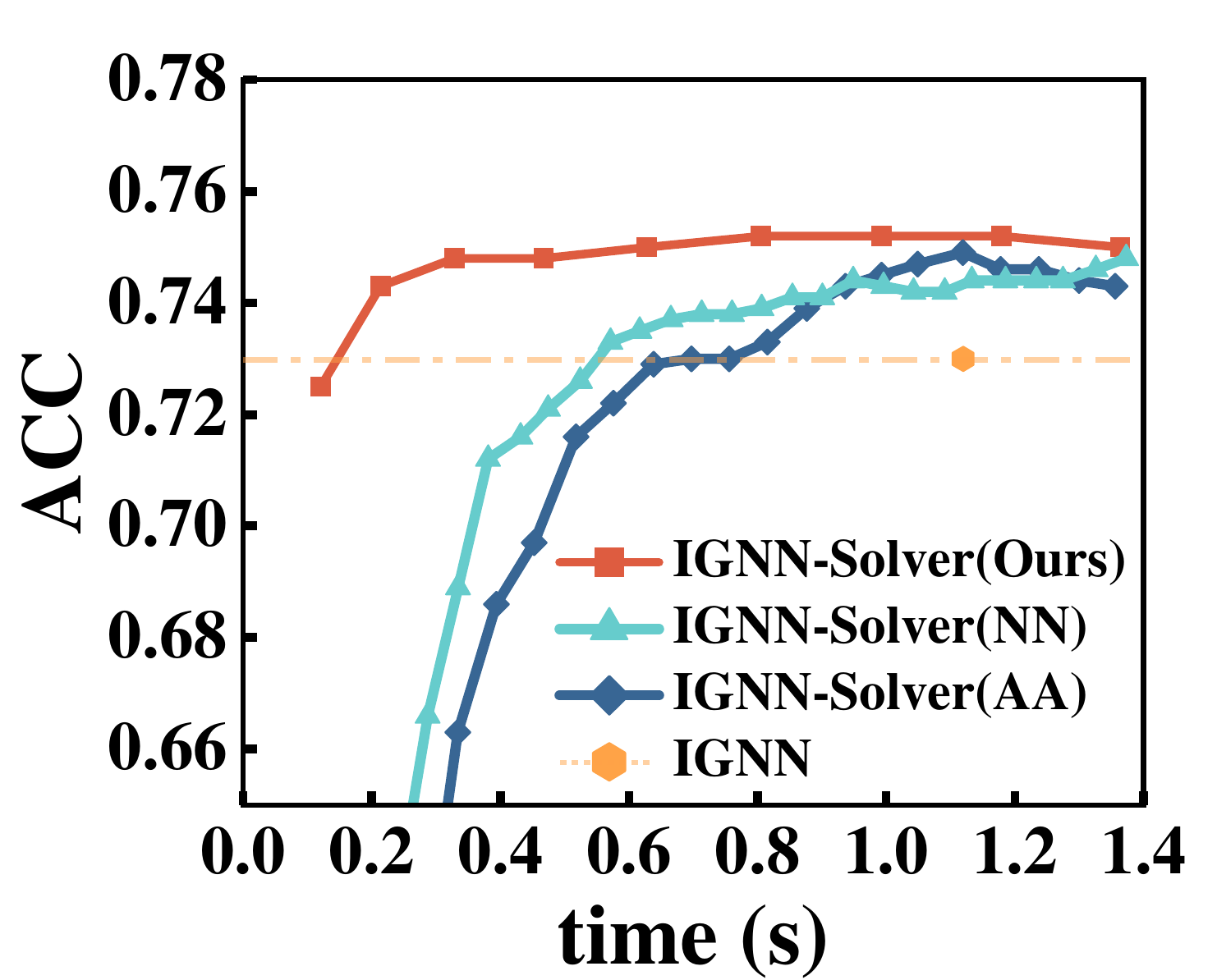}
            \caption{Citeseer}
            \label{t_Citeseer}
        \end{subfigure}
        \begin{subfigure}{3.4cm}
            \includegraphics[width=1\textwidth]{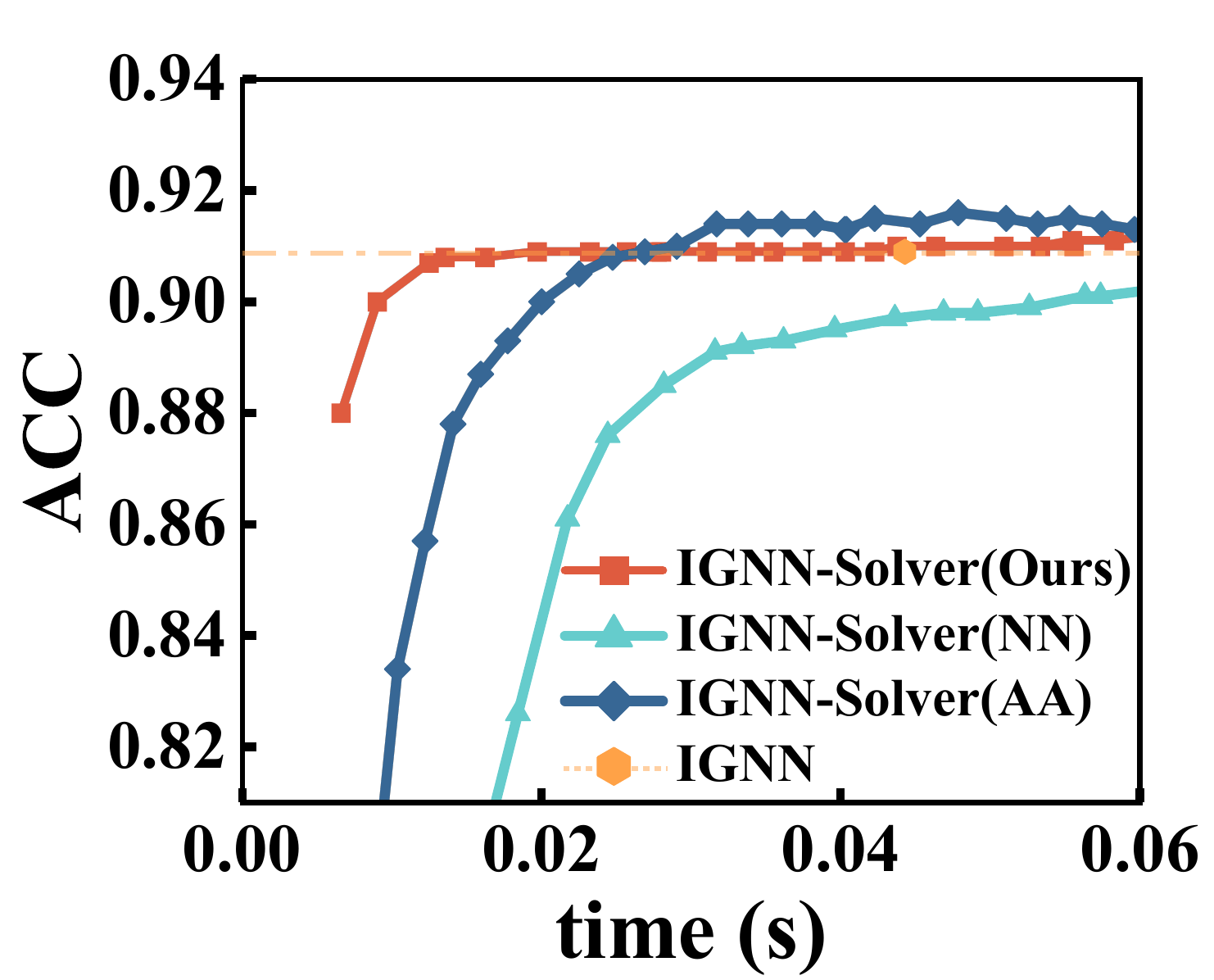}
            \caption{ACM}
            \label{t_ACM}
        \end{subfigure}
        \begin{subfigure}{3.4cm}
            \includegraphics[width=1\textwidth]{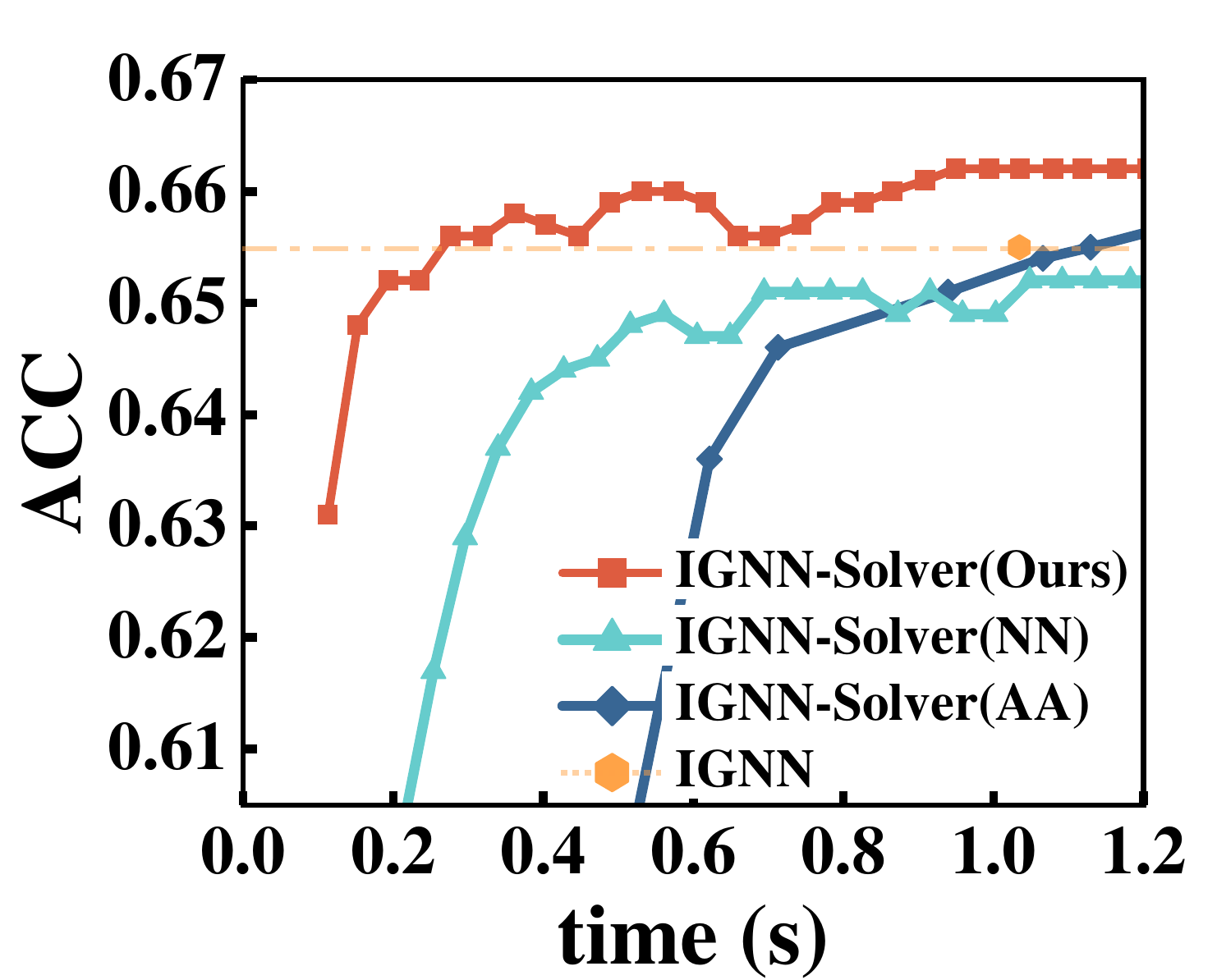}
            \caption{CoraFull}
            \label{t_CoraFull}
        \end{subfigure}
        \begin{subfigure}{3.4cm}
            \includegraphics[width=1\textwidth]{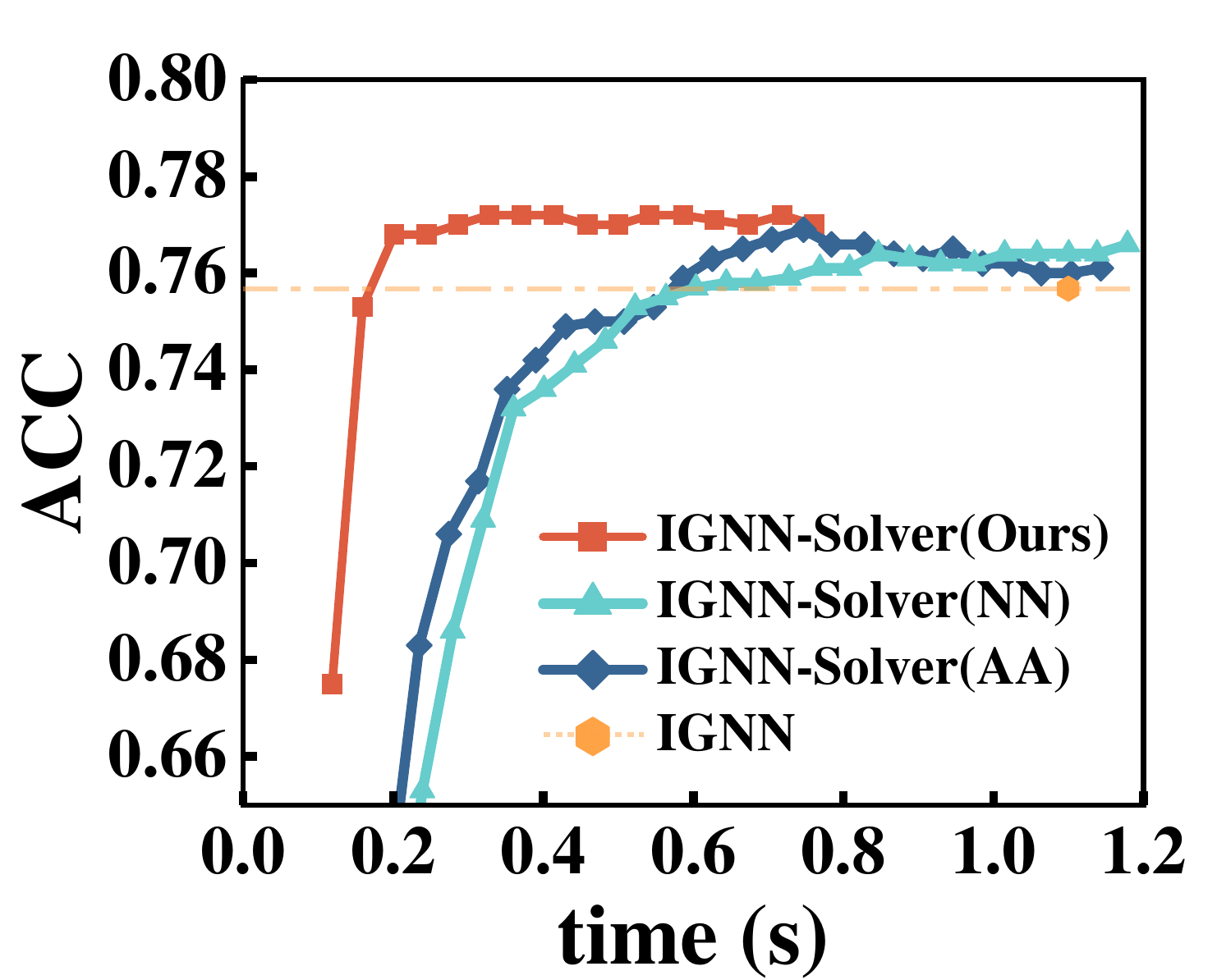}
            \caption{BlogCatalog}
            \label{t_BlogCatalog}
        \end{subfigure}
        \begin{subfigure}{3.4cm}
            \includegraphics[width=1\textwidth]{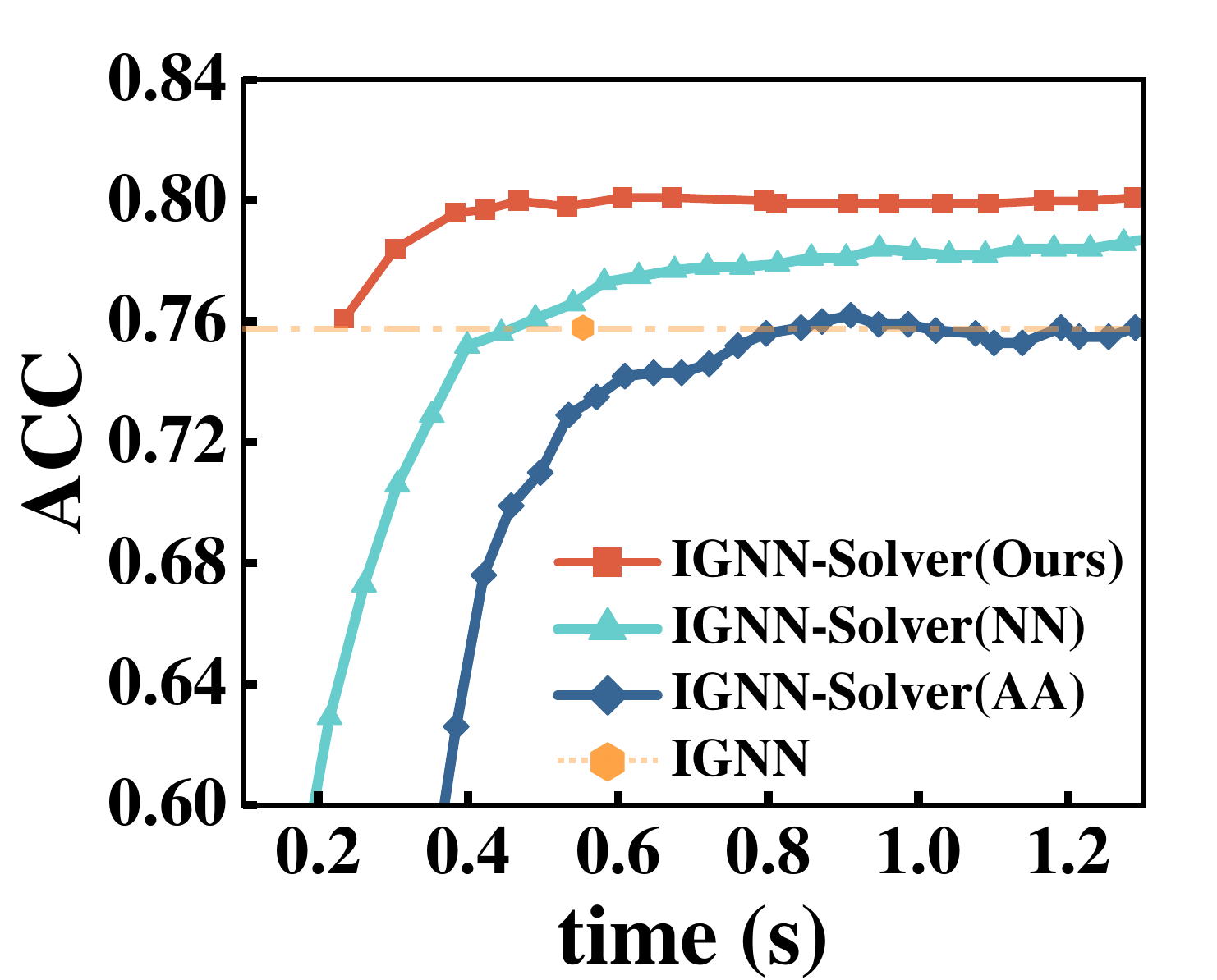}
            \caption{Flickr}
            \label{t_Flickr}
        \end{subfigure}

    \caption{Speed-accuracy tradeoff curves comparing the inference time of \textbf{\textcolor[RGB]{255, 162, 71}{IGNN}} with \textbf{\textcolor[RGB]{222, 92, 64}{IGNN-Solver} }and its variants \textbf{\textcolor[RGB]{102, 204, 204}{IGNN-Solver (NN)}} and \textbf{\textcolor[RGB]{56, 102, 148}{IGNN-Solver (AA)}}, across five \textit{small-scale} datasets: Citeseer~(\ref{t_Citeseer}), ACM~(\ref{t_ACM}), CoraFull~(\ref{t_CoraFull}), BlogCatalog~(\ref{t_BlogCatalog}) and Flickr~(\ref{t_Flickr}). Each plot represents the average performance over five independent runs under identical experimental conditions.
    }
    
    \label{F2_2}
\end{figure*}

\begin{table*}
\caption{Node Classification accuracy results on five \textit{small-scale} real-world datasets, with experiments conducted over five trials. The mean accuracy (\%) ± standard deviation is reported. The \textbf{best} and the \underline{runner-up} results are highlighted in boldface and underline, respectively.}
	\label{t1_2}
	\centering
		\begin{tabular}{clccccc}
			\toprule
			\multirow{3}{*}{Type}     & *Models                 & Citeseer                     & ACM                          & CoraFull                     & BlogCatalog                  & Flickr                       \\
			& *Nodes                  & $3,327$                      & $3,025$                      & $19,793$                     & $5,196$                      & $7,575$                      \\
			& *Edges                  & $4,732$                      & $13,128$                     & $65,311$                     & $171,743$                    & $239,738$                    \\ \midrule
			\multirow{9}{*}{Explicit} & GCN \cite{GCN}          & $64.80 \pm 4.15$             & $83.78 \pm 3.95$             & $35.98 \pm 9.44$             & $71.30 \pm 1.12$             & $76.98 \pm 1.83$             \\
			& GAT \cite{GAT}          & $71.24 \pm 2.88$             & $80.60 \pm 5.12$             & $50.24 \pm 1.87$             & $76.24 \pm 2.76$             & $72.64 \pm 3.23$             \\
			& SGC \cite{SGC}          & $70.32 \pm 2.75$             & $85.40 \pm 1.23$             & $46.56 \pm 4.43$             & $69.76 \pm 1.11$             & $71.88 \pm 3.47$             \\
			& APPNP \cite{APPNP}      & $61.56 \pm 8.92$             & $84.16 \pm 4.25$             & $21.24 \pm 4.76$             & $61.34 \pm 9.39$             & $73.42 \pm 3.80$             \\
			& JKNet \cite{JKNet}      & $63.78 \pm 8.76$             & $64.96 \pm 5.47$             & $23.04 \pm 6.01$             & $72.62 \pm 6.83$             & $75.68 \pm 1.15$             \\
			& AM-GCN \cite{AMGCN}     & $73.10 \pm 1.62$             & $89.56 \pm 0.30$             & $53.40 \pm 1.59$             & $73.86 \pm 1.10$             & $76.86 \pm 2.02$             \\
			& DEMO-Net \cite{demonet} & $68.34 \pm 2.94$             & $84.38\pm 2.19$              & $61.74 \pm 3.65$             & $74.26 \pm 2.70$             & $75.60\pm 3.95$              \\
			& GCNII  \cite{GCNII}     & $71.98 \pm 0.80$             & $85.36\pm  1.05$             & $57.64 \pm 3.34$             & $74.94 \pm 3.81$             & $ \underline{79.92}\pm 2.14$ \\
			& ACM-GCN \cite{ACM-GCN}  & $72.38 \pm 1.46$             & $88.98 \pm 0.41$             & $59.88 \pm 1.59$             & $\textbf{78.18}\pm 1.75$     & $74.82\pm 3.78$              \\ \midrule
			\multirow{6}{*}{Implicit} & IGNN \cite{IGNN}        & $72.96 \pm 1.83$             & $90.88 \pm 0.95$             & $65.52 \pm 0.51$             & $75.68 \pm 0.55$             & $75.80 \pm 0.29$             \\
			& EIGNN \cite{EIGNN}      & $72.38 \pm 1.36$             & $88.36 \pm 1.03$             & $61.80 \pm 0.60$             & $75.34 \pm 0.38$             & $75.66 \pm 0.94$             \\
			& MIGNN \cite{MIGNN}      & $73.79 \pm 0.94$             & $89.59 \pm 1.61$             & $62.94 \pm 0.46$             & $76.68 \pm 1.49$             & $74.96 \pm 0.49$             \\
			& IGNN-Solver (AA)     & $\underline{75.28} \pm 0.38$ & $\textbf{91.34} \pm 0.46$    & $\underline{65.88} \pm 0.34$ & $76.82 \pm 0.34$             & $75.84 \pm 0.27$             \\
			& IGNN-Solver (NN)     & $74.78 \pm 0.42$             & $91.08 \pm 0.56$             & $65.62 \pm 0.16$             & $76.88 \pm 0.74$             & $78.55 \pm 0.49$             \\
			& \textbf{IGNN-Solver (Ours)}  & $\textbf{75.60} \pm 0.22$    & $\underline{91.20} \pm 1.15$ & $\textbf{66.08} \pm 1.23$    & $\underline{77.64} \pm 0.54$ & $\textbf{80.14} \pm 0.23$    \\ \bottomrule
		\end{tabular}

\end{table*}

\subsection{Experiments Settings}
\subsubsection{Datasets}
The evaluation is conducted on nine different-field, real-world benchmarks for node classification tasks, including four citation datasets (Citeseer, ACM, CoraFull, ogbn-arxiv), three social interaction datasets (BlogCatalog, Flickr, Reddit), and two product network datasets (Amazon-all, ogbn-products); as well as five bioinformatics benchmarks (MUTAG, PTC\_MR, COX2, PROTEINS, NCI1) for graph classification tasks. Further details can be found in Appendix~\ref{secExperimental_Details}. Notably, to demonstrate the scalability of the proposed model on larger datasets, we use four \textit{large-scale datasets}: Amazon-all, Reddit, ogbn-arxiv, and ogbn-products, two of which are from the Open Graph Benchmark (OGB)~\cite{OGB}.

\subsubsection{Baselines}
We compare IGNN-Solver with a comprehensive set of baselines in graph node classification tasks,
including three implicit GNNs, i.e.,
MIGNN \cite{MIGNN},
EIGNN \cite{EIGNN},
IGNN \cite{IGNN},
and nine explicit/traditional GNNs, i.e.,
AM-GCN \cite{AMGCN},
GCN \cite{GCN},
GAT \cite{GAT},
SGC \cite{SGC},
APPNP \cite{APPNP},
JKNet \cite{JKNet},
DEMO-Net \cite{demonet},
GCNII  \cite{GCNII},
ACM-GCN \cite{ACM-GCN}.
For the sake of fairness, we use the same hidden layer configuration for all baseline models. For example, in the Flickr dataset, the dimension of the hidden layers is uniformly set to $512$ for all methods (including ours, see in Table~\ref{Hp_setting}).
In addition, for EIGNN, the arbitrary gamma is set to $0.8$, which is consistent with the original paper.
For AM-GCN, the number of nearest neighbors in the KNN graph is set from {$5,6,7$}.
For GCN, the hidden layer setup is the same as that for others.
For GAT, three attention headers are used for each layer.
For SGC, the power of self-loops in the graph adjacency matrix is set to $3$.
For APPNP, we set the number of iterations to $3$ and teleport probability $\alpha$ to $0.5$.
For JKNet, we set layers to $1$ in small datasets and set to $8$ in large ones.
For DEMO-Net, the regularization parameter is set as $5e-4$, the learning rate is set as $5e-3$, and the hash dimension is set as $256$.
For GCNII, we set $\alpha_{\ell}=0.1$ and $L_2$ regularization to $5e-4$, consistent with the original paper.
For ACM-GCN, we set layers to $1$ in small datasets and set to $4$ in large ones.
For our IGNN-Solver and its variants, we employ the same settings as the basic IGNN model.
Furthermore, we adjust these parameters affecting the convergence and performance of IGNN through a combined approach of hierarchical grid search and manual tuning. The Adam optimizer~\cite{Adam} is used for optimization.


\subsection{Performance on Node Classification Tasks}\label{sec51}

\subsubsection{Experimental setup}
To demonstrate the superiority of the IGNN-Solver over IGNNs in terms of both performance and efficiency, we analyze the movement of the speed/accuracy Pareto curve across various datasets, rather than concentrating on a single point on the curve, which is depicted in Figures ~\ref{F2} and~\ref{F2_2}. All experiments are conducted five times, and the best results are reported. 
Furthermore, we present the node classification performance results for each method on various datasets, which are shown in Tables~\ref{t1_1} and~\ref{t1_2}. All experiments are conducted five times, and we report the average results along with the standard deviation.
Additional details on the training procedure and the hyperparameters used for each task are provided in Appendices~\ref{secExperimental_Details} and~\ref{Appendix_Training_Strategies}.

\subsubsection{Results on large-scale tasks}
From Figure~\ref{F2} (the speed/accuracy Pareto curve on large-scale datasets: Amazon-all, Reddit, ogbn-arxiv and ogbn-products), we observe that:
\begin{enumerate}
    \item \textit{Comprehensive performance and efficiency}: IGNN-Solver generally outperforms all other methods, particularly in large datasets with a greater graph radius. This advantage is more pronounced in such cases. The improved performance is attributed to IGNN-Solver’s significant enhancement in the convergence path and the better initialization of the fixed-point equation through the initializer;
    \item \textit{Rapid inference advantage}: IGNN-Solver demonstrates a notable speed advantage during inference. For instance, on the large-scale dataset Amazon-all, our IGNN-Solver achieves $8\times$ faster inference speed than IGNN while maintaining the same performance, and there is consistently at least $1.5\times$ acceleration and a similar pattern has been observed as well on other datasets.
\end{enumerate}

From Table~\ref{t1_1}, we observe that implicit GNNs with infinite depth generally outperform shallow explicit GNNs in most cases. Furthermore, IGNN-Solver consistently shows higher accuracy percentages across most datasets compared to other state-of-the-art explicit GNNs like DEMO-Net~\cite{demonet}, GCNII~\cite{GCNII} etc, indicating its superior performance.


\subsubsection{Results on small-scale tasks}
We further evaluate IGNN-Solver on several small-scale graph node classification tasks, including Citeseer, ACM, CoraFull, BlogCatalog, and Flickr. The hyperparameters used are outlined in Table~\ref{Hp_setting}, Section~\ref{sec_Hp_setting}. The mean accuracy and standard deviation are reported in Table~\ref{t1_2}, while the speed/accuracy Pareto curve is presented in Figure~\ref{F2_2}.
In general, learning LRD is not crucial for these tasks, as the graph diameters are relatively small~\cite{MIGNN}. However, as shown in Table~\ref{t1_2}, even for these small-scale node classification tasks, IGNN-Solver outperforms IGNNs and many explicit GNNs, as well as other advanced implicit models. As illustrated in Figure~\ref{F2_2}, the solver continues to improve the convergence path of IGNN, even in small-scale tasks.
These results, along with those in Section~\ref{sec51}, confirm the expressive power of the IGNN-Solver using learnable graph neural networks, even surpassing that of many explicit and enhanced implicit GNNs.

\subsection{Performance on Graph Classification Tasks}
\label{sec_graph_classification}
\begin{table*}
    \centering
    \caption{Graph classification accuracy results on five benchmark datasets. The mean accuracy (\%) ± standard deviation is reported. The \textbf{best} and the \underline{runner-up} results are highlighted in boldface and underline, respectively.}
    \begin{tabular}{llccccc}
        \toprule
        Type & Method & MUTAG & PTC\_MR & COX2 & PROTEINS & NCI1 \\
        \midrule
        \multirow{7}{*}{Explicit}
        & WL~\cite{WL} & 84.1 $\pm$ 1.9 & 58.0 $\pm$ 2.5 & 83.2 $\pm$ 0.2 & 74.7 $\pm$ 0.5 & \textbf{84.5 $\pm$ 0.5} \\
        & DCNN~\cite{DCNN} & 67.0 & 56.6 & - & 61.3 & 62.6 \\
        & DGCNN~\cite{DGCNN} & 85.8 & 58.6 & - & 75.5 & 74.4 \\
        & GIN~\cite{GIN} & \underline{89.4 $\pm$ 5.6} & 64.6 $\pm$ 7.0 & - & 76.2 $\pm$ 2.8 & \underline{82.7 $\pm$ 1.7} \\
        & FDGNN~\cite{FDGNN} & 88.5 $\pm$ 3.8 & 63.4 $\pm$ 5.4 & 83.3 $\pm$ 2.9 & 76.8 $\pm$ 2.9 & 77.8 $\pm$ 1.6 \\
        & PK~\cite{PK} & 76.0 $\pm$ 2.7 & 59.5 $\pm$ 2.4 & 81.0 $\pm$ 0.2 & 73.7 $\pm$ 0.7 & 82.5 $\pm$ 0.5 \\
        & GCN~\cite{GCN} & 85.6 $\pm$ 5.8 & 64.2 $\pm$ 4.3 & - & 76.0 $\pm$ 3.2 & 80.2 $\pm$ 2.0 \\
        \midrule
        \multirow{5}{*}{Implicit}
        & IGNN~\cite{IGNN} & 76.0 $\pm$ 13.4 & 60.5 $\pm$ 6.4 & 79.7 $\pm$ 3.4 & 76.5 $\pm$ 3.4 & 73.5 $\pm$ 1.9 \\
        & CGS~\cite{CGS} & \underline{89.4 $\pm$ 5.6} & 64.7 $\pm$ 6.4 & - & 76.3 $\pm$ 4.9 & 77.6 $\pm$ 2.0 \\
        & GIND~\cite{GIND} & 89.3 $\pm$ 7.4 & 66.9 $\pm$ 6.6 & 84.8 $\pm$ 4.2 & 77.2 $\pm$ 2.9 & 78.8 $\pm$ 1.7 \\
        & MIGNN~\cite{MIGNN} & 81.8 $\pm$ 9.1 & \textbf{72.6 $\pm$ 5.4} & \underline{85.0 $\pm$ 5.3} & \underline{77.9 $\pm$ 2.6} & 79.6 $\pm$ 1.8 \\
		&\textbf{IGNN-Solver (Ours)} & \textbf{90.5 $\pm$ 1.6} & \underline{71.8 $\pm$ 3.1} & \textbf{85.3 $\pm$ 3.0} & \textbf{78.4 $\pm$ 2.6} & {80.2 $\pm$ 1.1}\\
        \bottomrule
    \end{tabular}
    \label{tab:graph_classification_results}
\end{table*}

\subsubsection{Experimental setup}
We extend our evaluation to graph classification tasks to further demonstrate the effectiveness and versatility of IGNN-Solver. We conduct experiments on several widely used bioinformatics benchmark datasets for graph classification, including MUTAG, PTC\_MR, COX2, PROTEINS, and NCI1. These datasets vary in size and complexity, as detailed in Appendix~\ref{secExperimental_Details}. 
We train the model using 10-fold cross-validation using the experimental setup of~\cite{IGNN}. For each dataset, we employ standard 10-fold cross-validation to evaluate the test performance of the IGNN-Solver, with the baseline and their results drawn from~\cite{MIGNN, GIND}. The average prediction accuracy and standard deviations are in Table~\ref{tab:graph_classification_results}.

\subsubsection{Results and discussion}
The results of the graph classification experiments, summarized in Table~\ref{tab:graph_classification_results}, show that our IGNN-Solver almost outperforms the baseline models across all datasets, clearly demonstrating its ability to effectively capture graph structures and significantly improve classification accuracy. These findings confirm the effectiveness of the IGNN-Solver in graph classification tasks, further supporting its potential for deployment in real-world applications where graph data is prevalent.

\subsection{Empirical Understandings of IGNN-Solver}\label{sec_others}
\subsubsection{Overhead in training}\label{sec_model_size}

\begin{figure}
\centering
\begin{subfigure}{6.5cm}
    \includegraphics[width=1\textwidth]{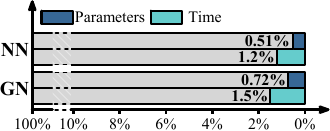}
    \caption{Overhead proportion on Citeseer.}
    \label{GNNN_ci}
\end{subfigure}
\begin{subfigure}{6.5cm}
    \includegraphics[width=1\textwidth]{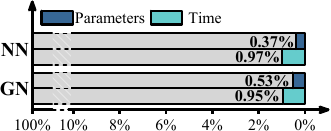}
    \caption{Overhead proportion on Amazon-all.}
    \label{GNNN_am}
\end{subfigure}
\caption{Relative parameter size and training time of the IGNN-Solver on the small Citeseer dataset and the large Amazon-All dataset, with consistent patterns observed across other datasets.}
\label{f_Modelsize}
\end{figure}

To investigate the proportion of training overhead attributed to the solver throughout the entire training process, we depict in Figure~\ref{f_Modelsize} the percentage of training computations on IGNN-Solver relative to the total time overhead on IGNN. Our approach is not only effective but also incurs minimal training overhead for the solver: the solver module is relatively small and requires only about $1\%$ of the original training time required by the IGNN model.
Besides, this proportion will be lower for large-scale data. For instance, on the Amazon dataset, IGNN necessitates a total runtime of $3$ hours, whereas the solver only requires approximately $1.6$ minutes, indicating that our solver maintains its lightweight nature across any-scale datasets.

\subsubsection{Efficiency study}

\begin{figure*}
    \centering
    \begin{subfigure}{0.24\textwidth}
        \includegraphics[height=3.2cm]{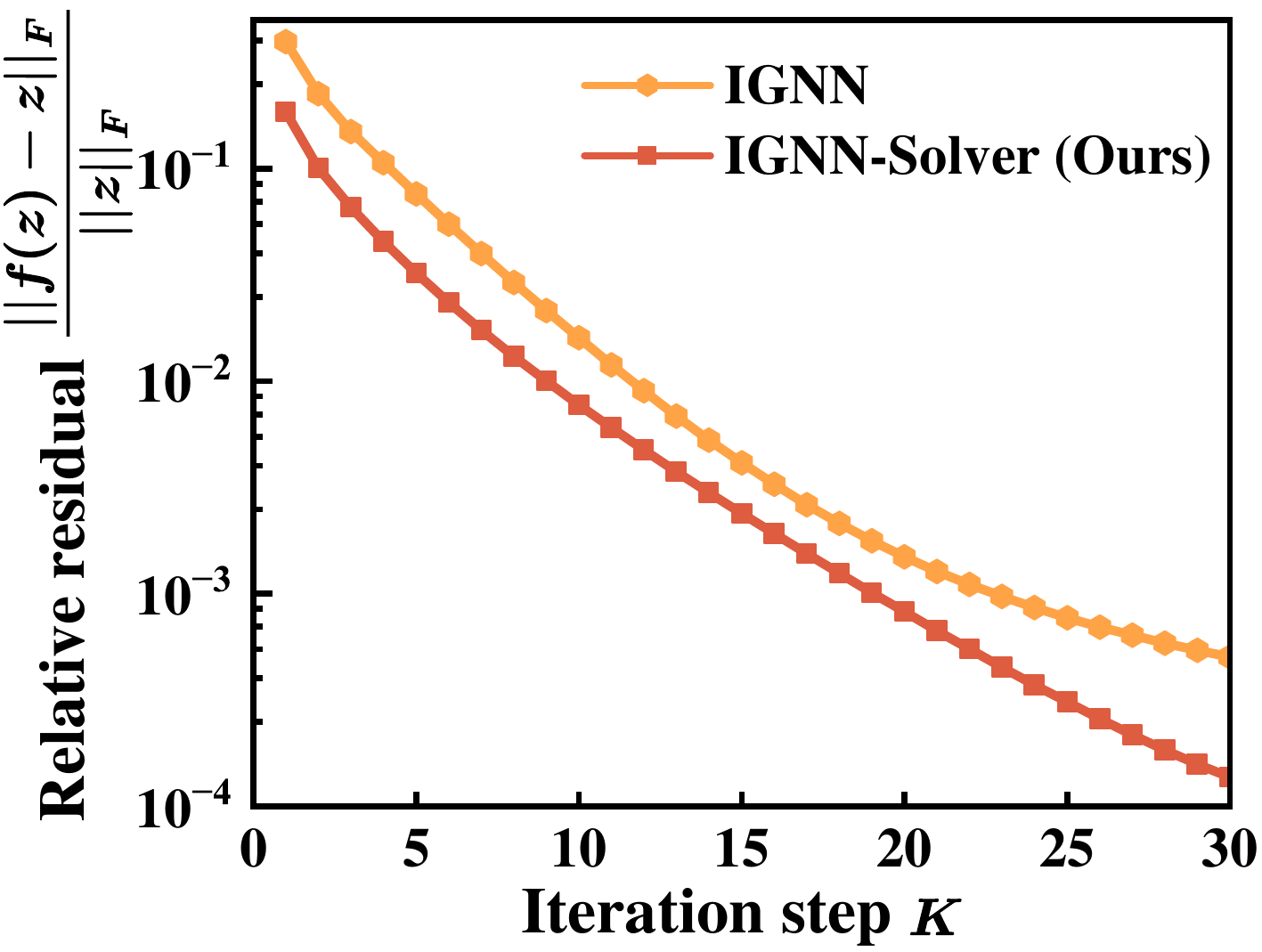}
        \caption{Amazon-all}
    \end{subfigure}
    \begin{subfigure}{0.24\textwidth}
        \includegraphics[height=3.2cm]{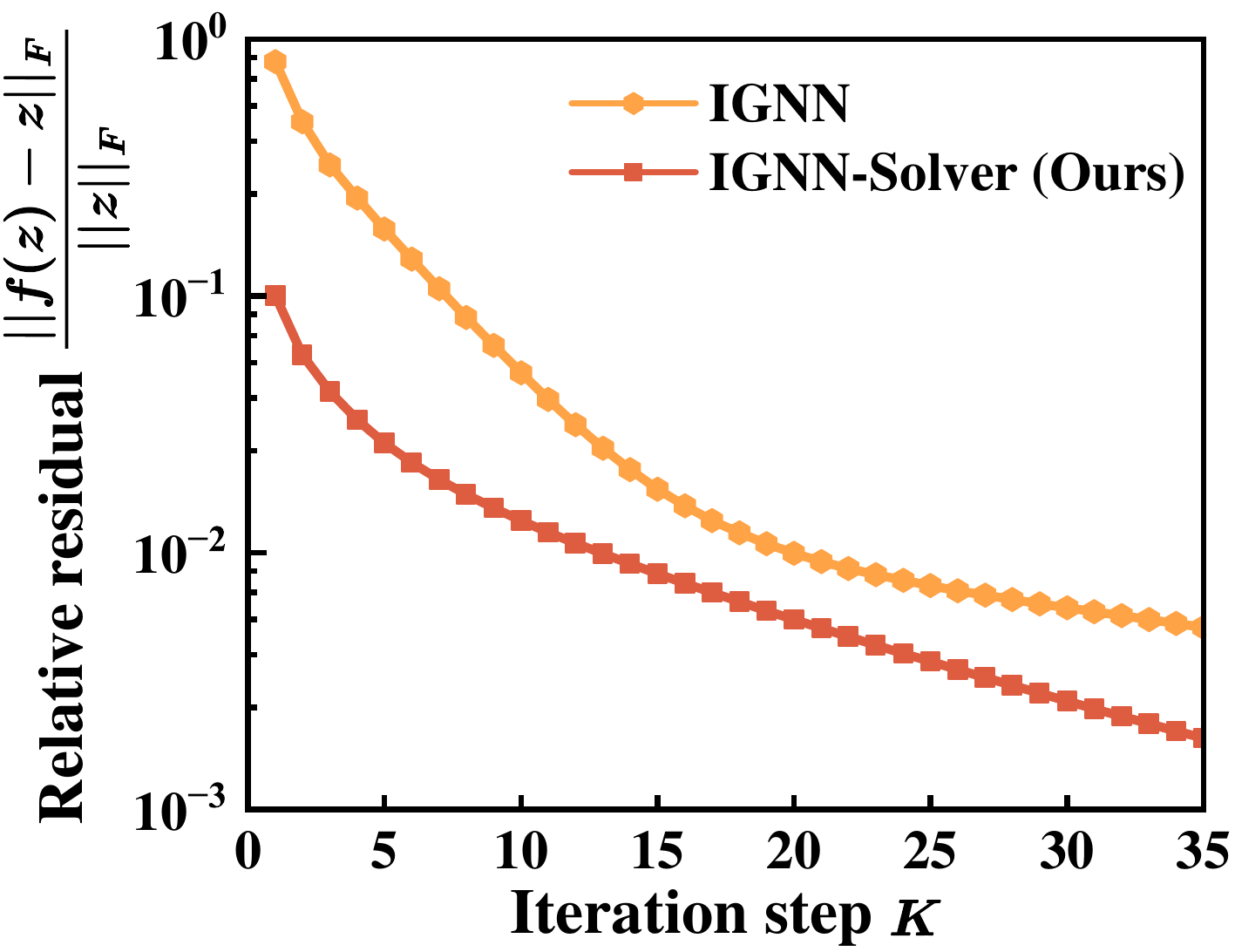}
        \caption{Reddit}
    \end{subfigure}
    \begin{subfigure}{0.24\textwidth}
        \includegraphics[height=3.2cm]{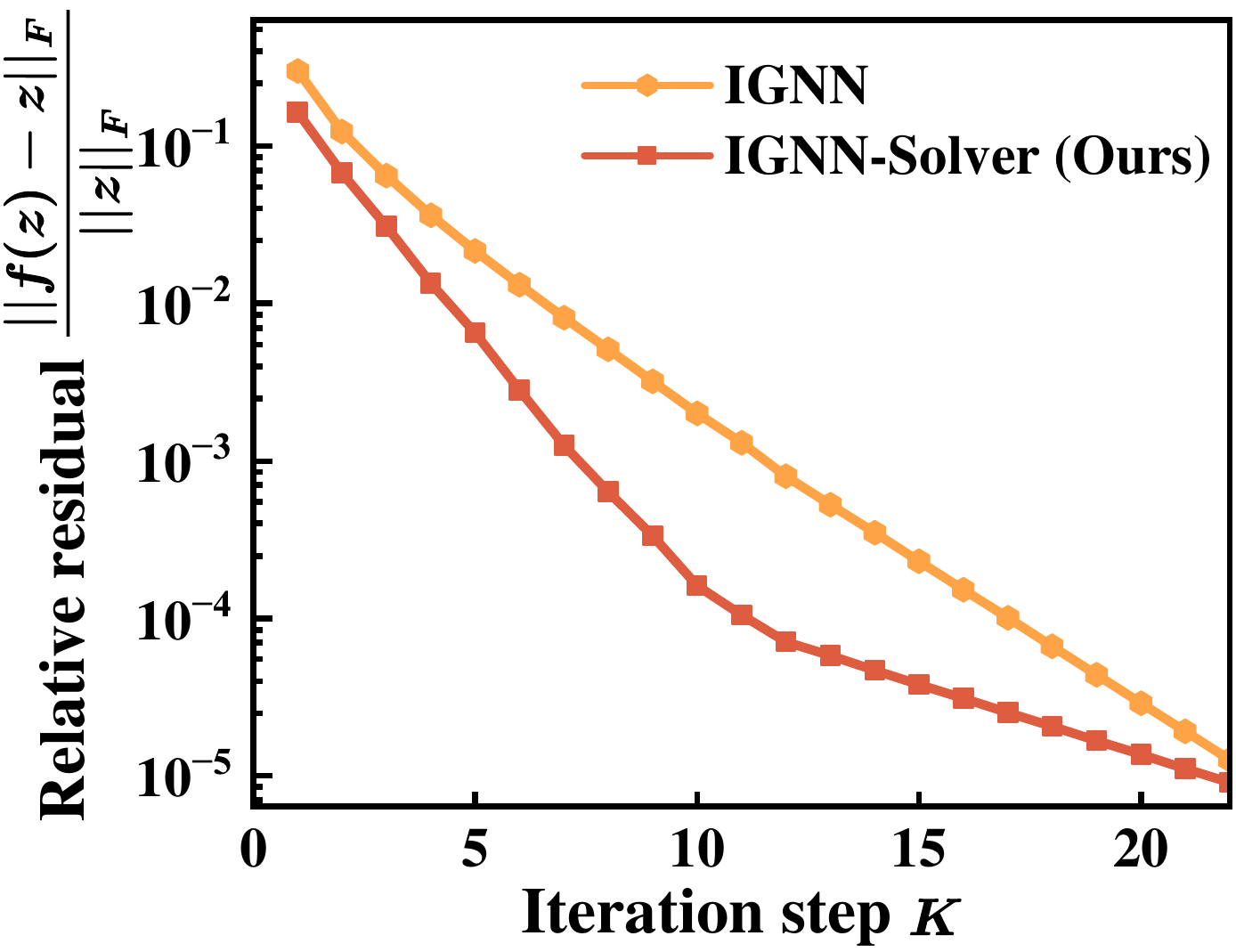}
        \caption{obgn-arxiv}
    \end{subfigure}
    \begin{subfigure}{0.24\textwidth}
        \includegraphics[height=3.2cm]{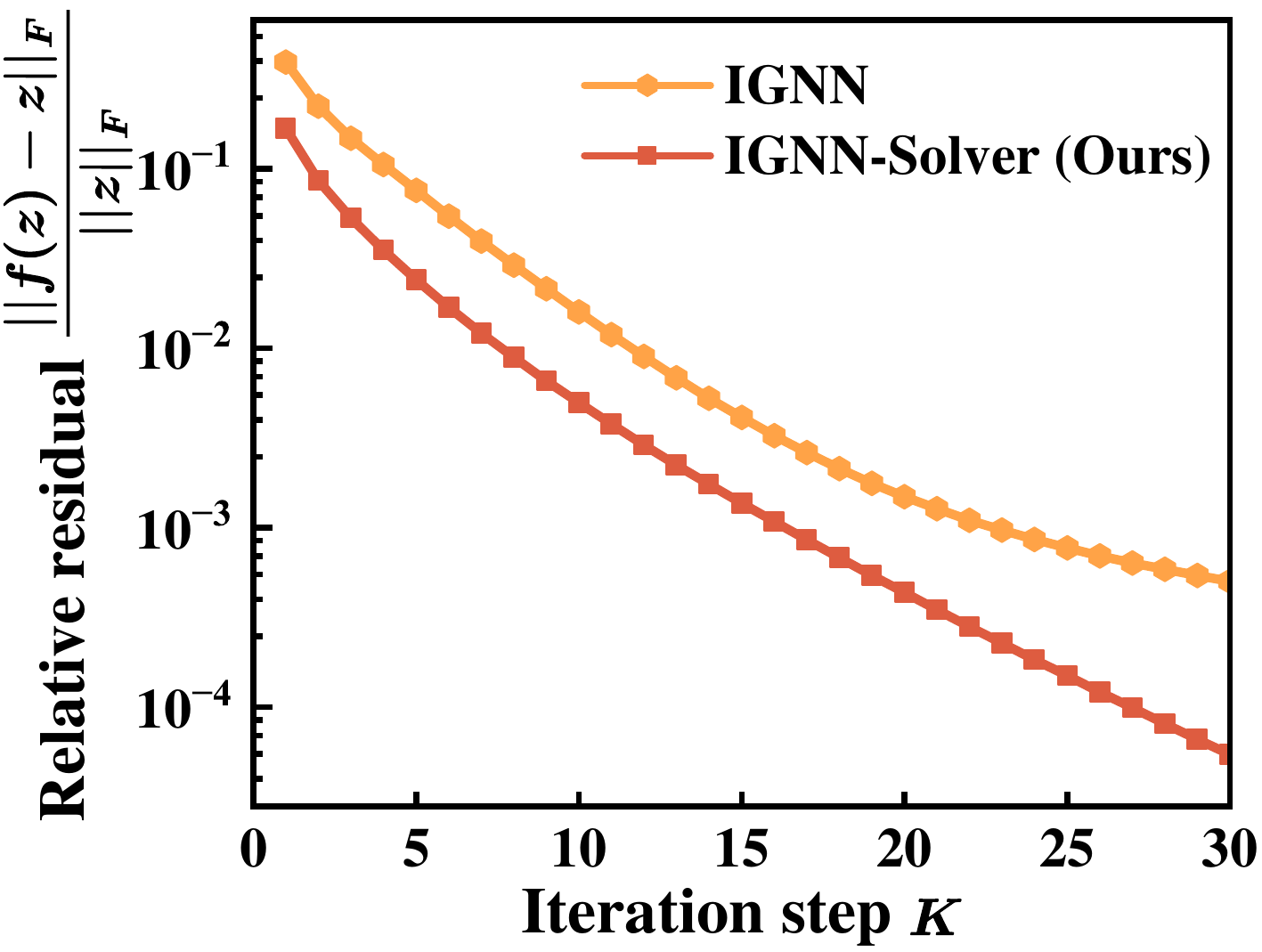}
        \caption{obgn-products}
    \end{subfigure}
    \begin{subfigure}{0.24\textwidth}
        \includegraphics[height=3.2cm]{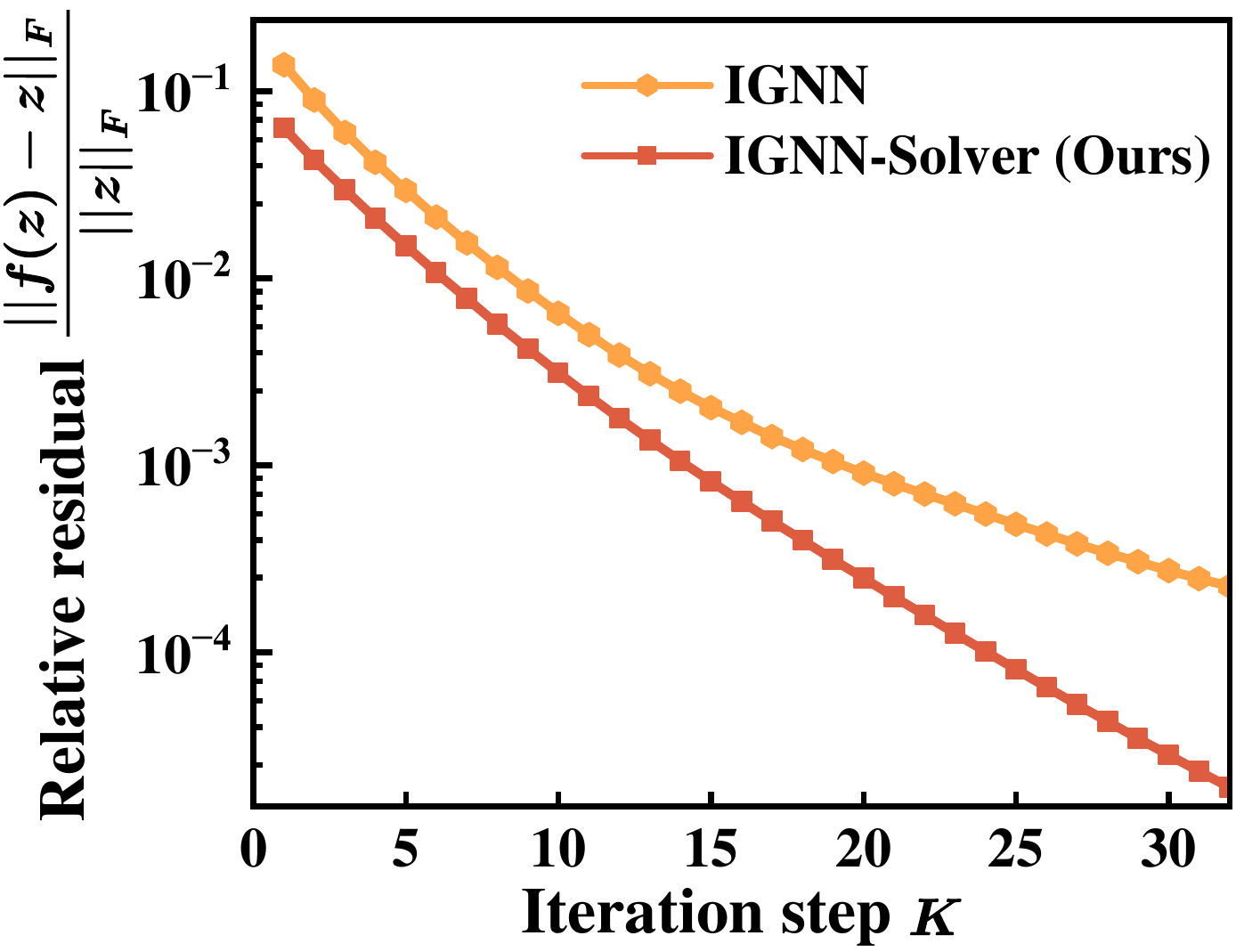}
        \caption{MUTAG}
    \end{subfigure}
    \begin{subfigure}{0.24\textwidth}
        \includegraphics[height=3.2cm]{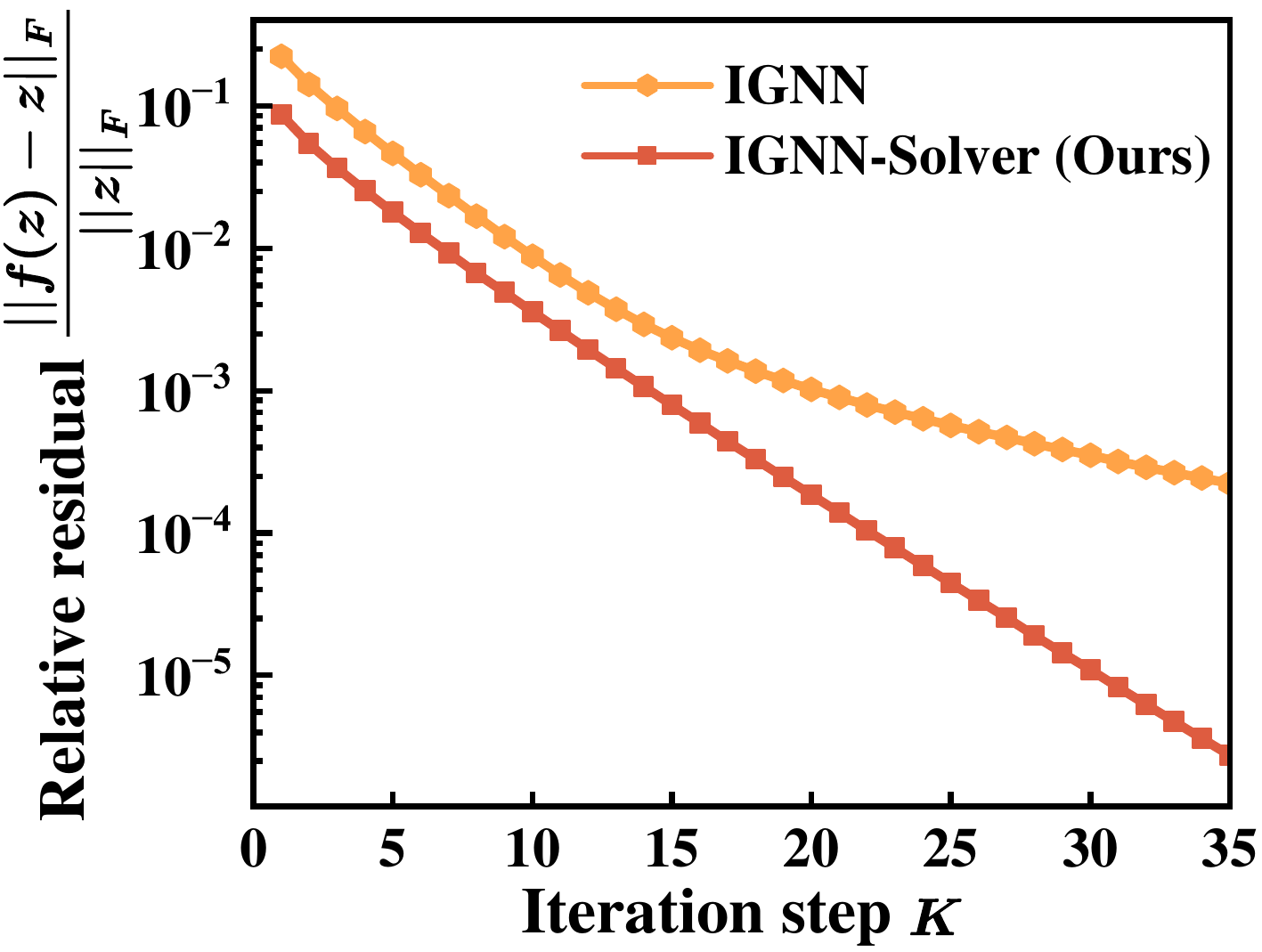}
        \caption{PTC\_MR}
    \end{subfigure}
    \begin{subfigure}{0.24\textwidth}
        \includegraphics[height=3.2cm]{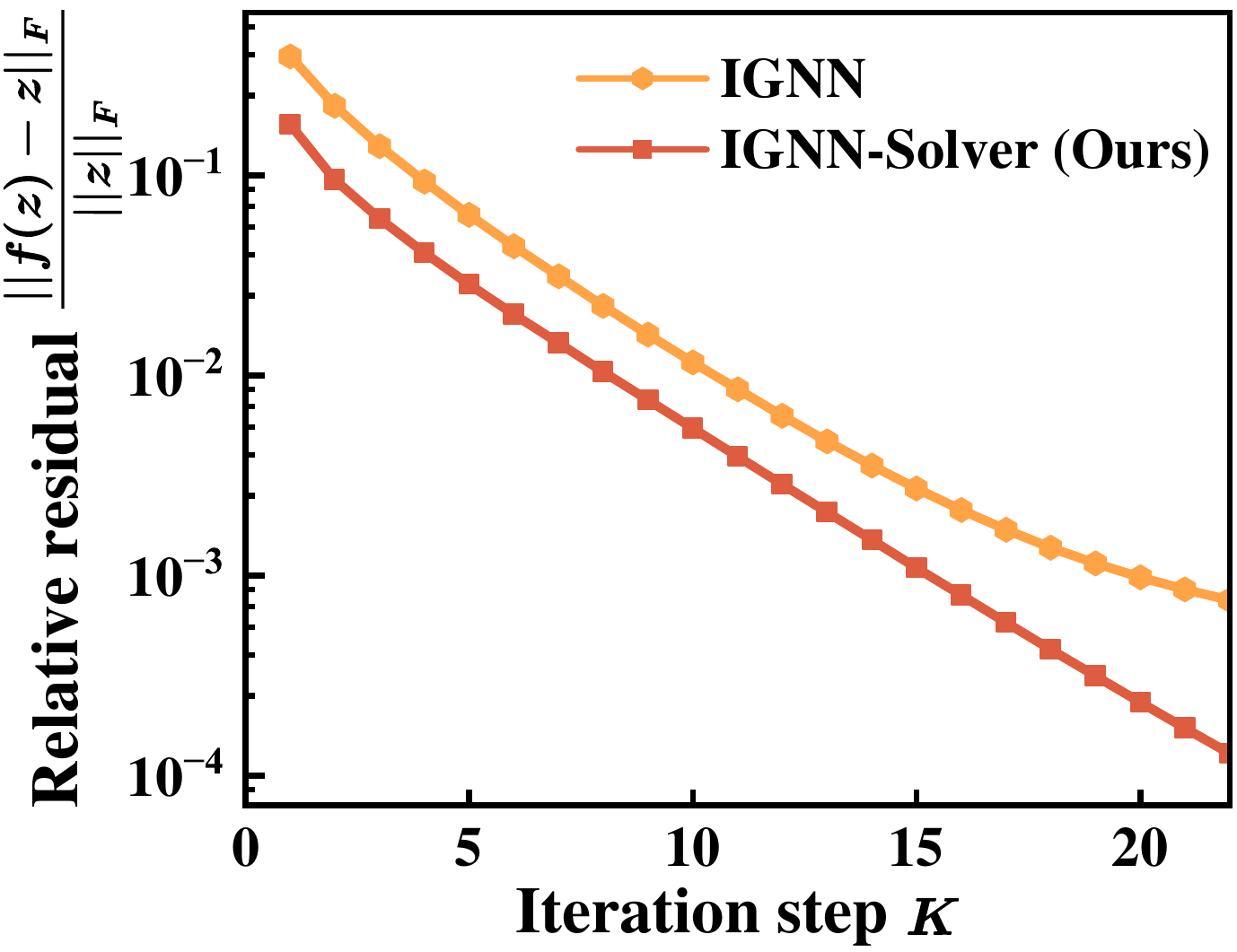}
        \caption{COX2}
    \end{subfigure}
    \begin{subfigure}{0.24\textwidth}
        \includegraphics[height=3.2cm]{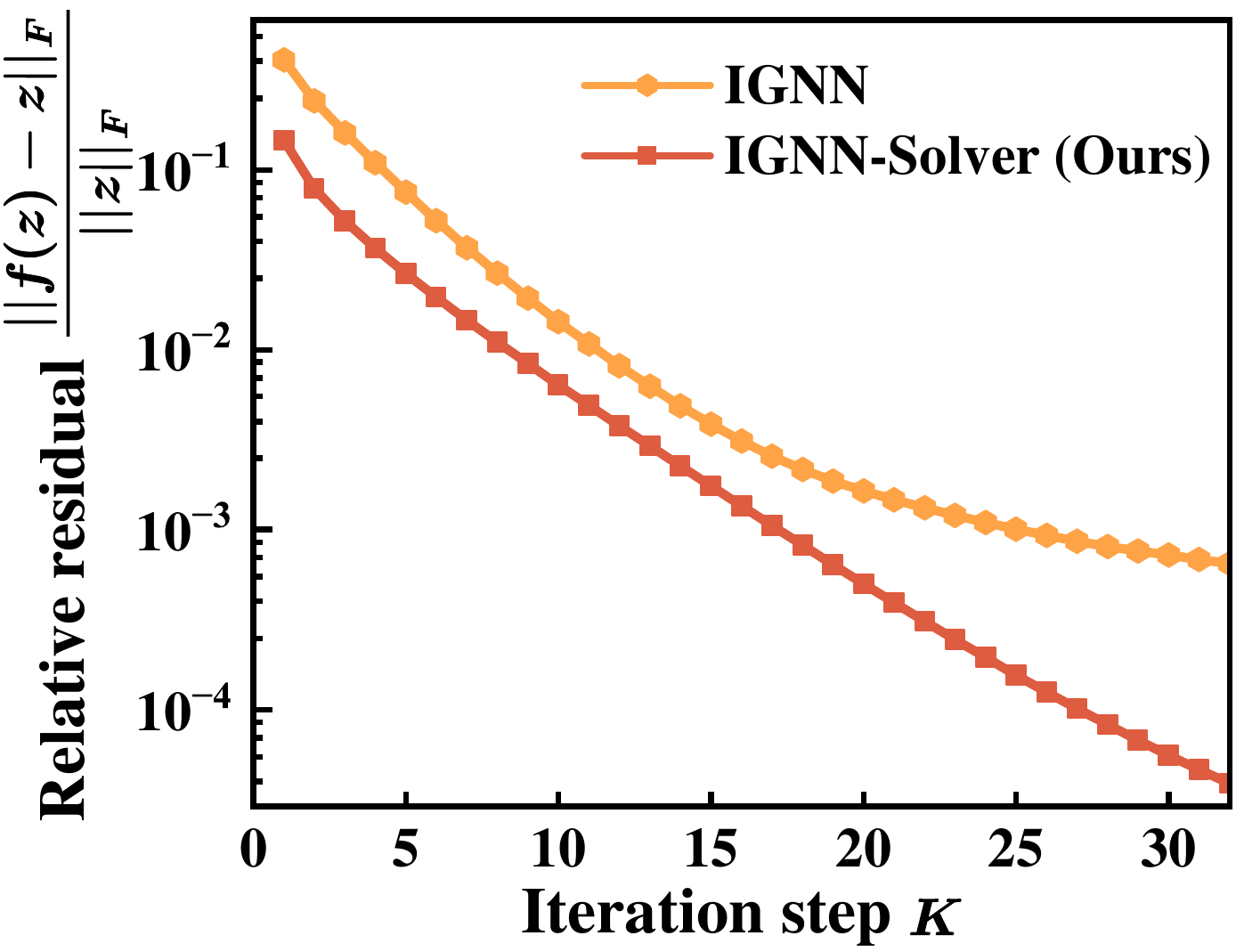}
        \caption{PROTEINS}
    \end{subfigure}
    \caption{
    Comparison of convergence curves across eight datasets, highlighting the substantial acceleration in convergence achieved by our IGNN-Solver.
    }
    \label{rel_diff}
\end{figure*}

We present additional evidence on the convergence and generalizability of the neural solvers in Figure~\ref{rel_diff}, where the ordinate represents the relative residual $\frac{||f(z)-z||_F}{||z||_F}$, with $\|\cdot\|_F$ denoting the Frobenius norm. We compare the convergence of a pre-trained IGNN model under two conditions: 1) canonical iteration in IGNN~\cite{IGNN}, and 2) IGNN-Solver with $K = 6$ unrolling steps. 

From Figure~\ref{rel_diff}, we observe that the IGNN-Solver, trained with $K$ unrolling steps, is capable of generalizing beyond $K$, with both solvers ultimately reaching a stable state, demonstrating good convergence. Moreover, thanks to the fixed-point solution of the graph neural parameterization, IGNN-Solver continuously enhances the convergence of the canonical iterators while being more computationally efficient. Specifically, each step of the neural solver is cheaper than the standard iterator step. This explains the observed improvement in inference efficiency of at least $1.5\times$ in Section~\ref{sec51}.

\subsubsection{Other choices for graph neural layers in IGNN-Solver}\label{sec_curve}

\begin{figure}
    \centering
    \begin{subfigure}{6cm}
        \includegraphics[width=1.0\textwidth]{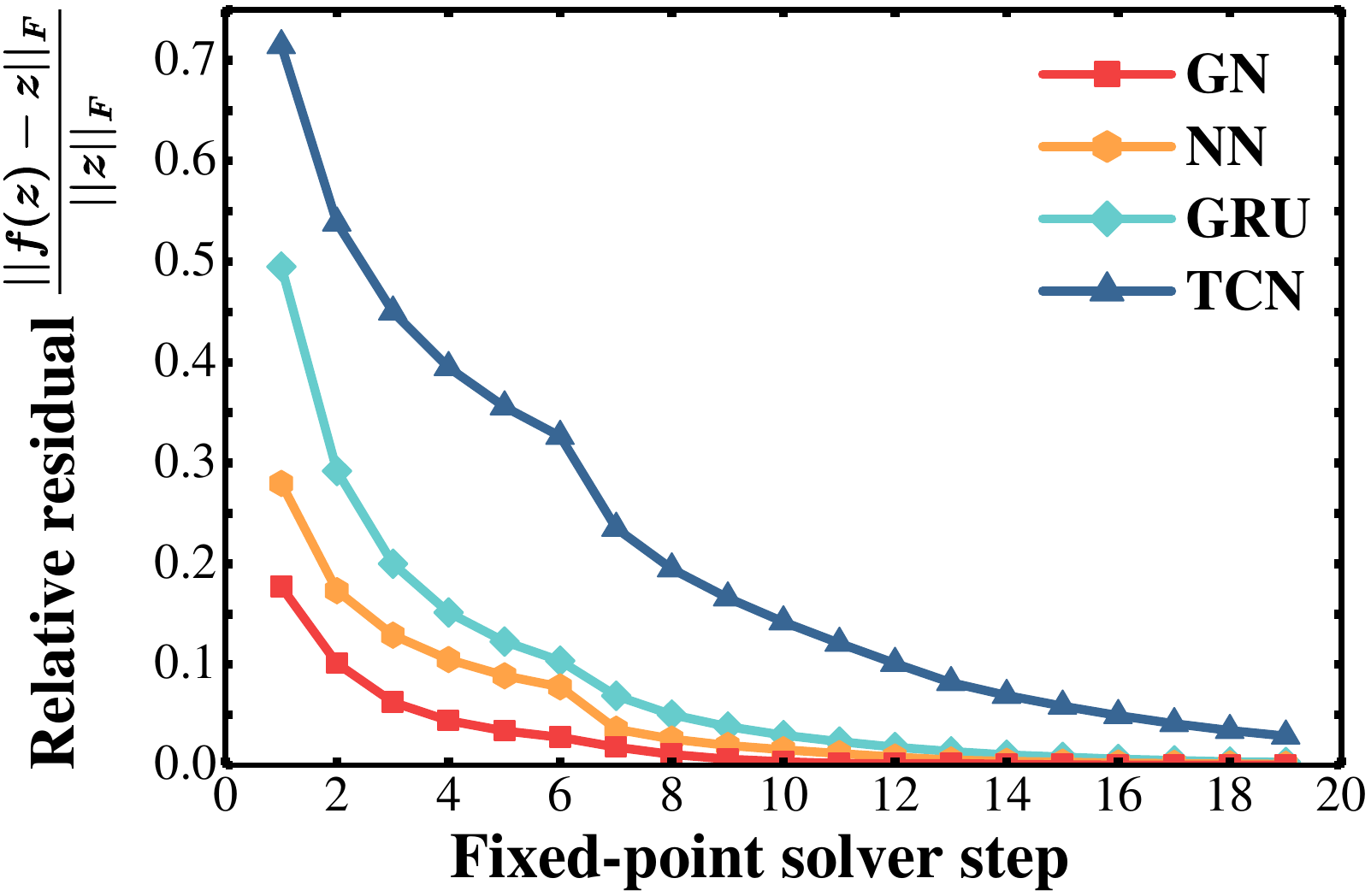}
        \caption{The relative error curve during warm-up.}
    \end{subfigure}
    \begin{subfigure}{6cm}
        \includegraphics[width=1.0\textwidth]{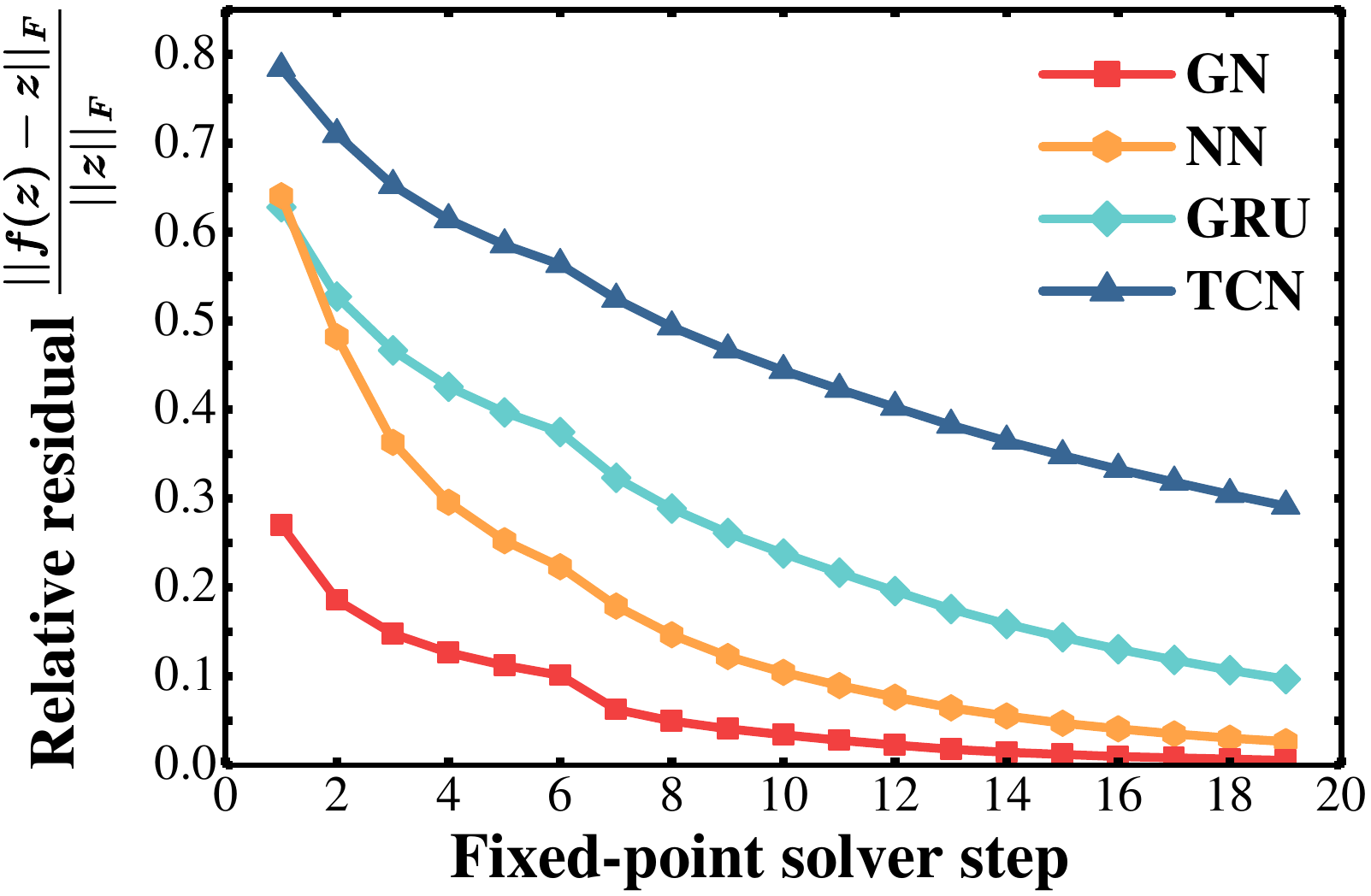}
        \caption{The relative error curve during inference.}
    \end{subfigure}
    \caption{
    Comparison of IGNN-Solver with \textcolor[RGB]{242, 64, 64}{\textbf{GN}}, \textcolor[RGB]{255, 163, 71}{\textbf{NN}}, \textcolor[RGB]{102, 204, 204}{\textbf{GRU}}, and \textcolor[RGB]{56, 102, 148}{\textbf{TCN}}. Our proposed IGNN-Solver (with \textcolor[RGB]{242, 64, 64}{\textbf{GN}}) achieves the fastest convergence and superior overall performance, owing to its lightweight design and ability to effectively utilize rich graph information.
    }
    \label{Curves_R}
\end{figure}

We provide additional evidence on the convergence and generalization of IGNN-Solver by exploring alternatives to our proposed graph neural (GN) layers, including neural network (NN), temporal convolutional network (TCN)~\cite{TCN}, and gated recurrent unit (GRU)~\cite{GRU}. The relative residual curves on the Citeseer dataset, observed during both the warm-up and final inference stages, are presented in Figure~\ref{Curves_R}. Interestingly, despite their larger number of parameters and more complex architectures, TCN and GRU consistently underperform compared to NN. Notably, IGNN-Solver (GN) significantly improves the convergence path of all solvers and exhibits the fastest rate of residual reduction, establishing itself as the most effective predictor in IGNN-Solver due to its ability to effectively exploit rich graph information.


\begin{figure*}
    \centering
    \begin{subfigure}{0.3\textwidth}
        \includegraphics[height=4.5cm]{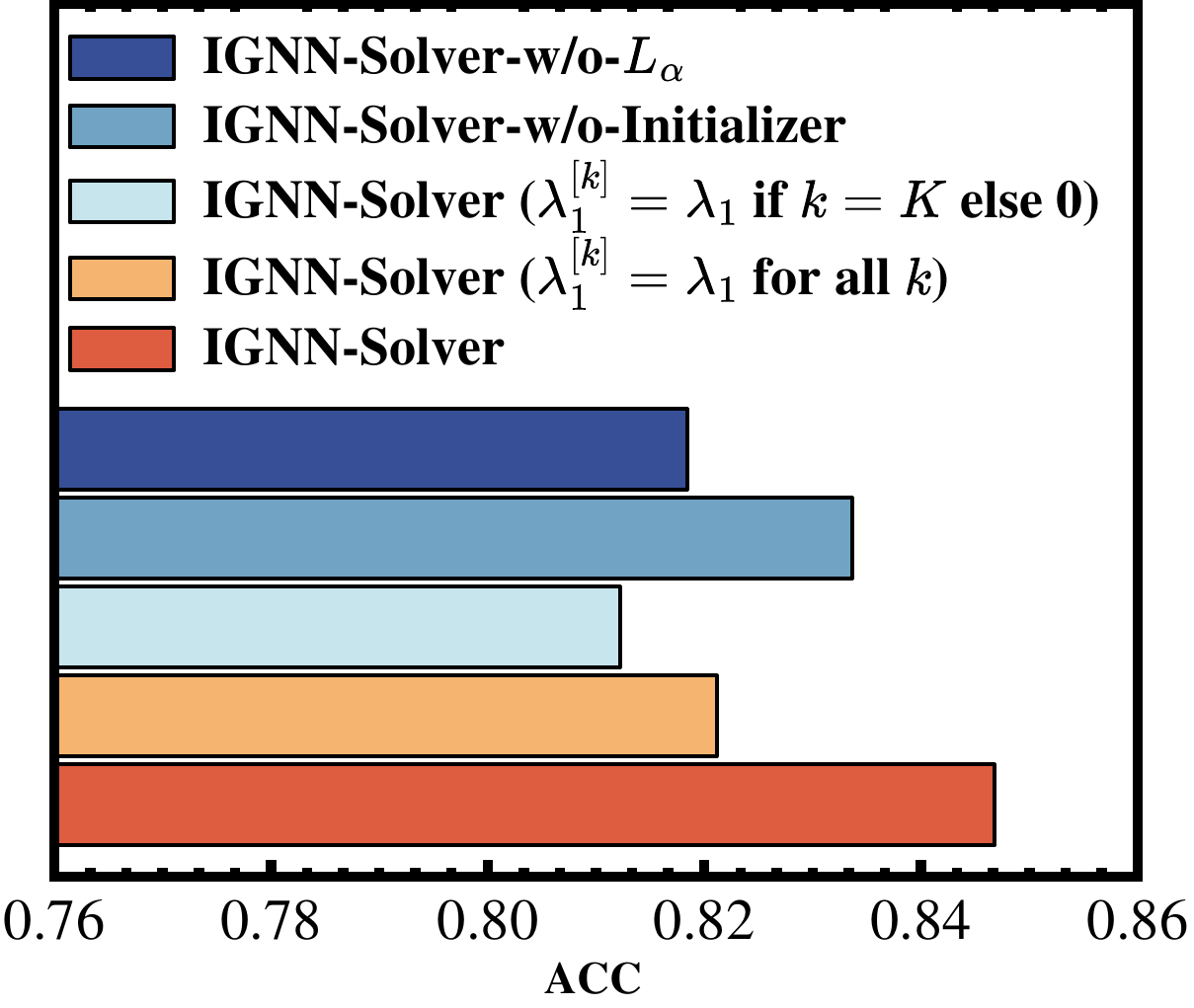}
        \caption{Amazon-all}
    \end{subfigure}
    \begin{subfigure}{0.3\textwidth}
        \includegraphics[height=4.5cm]{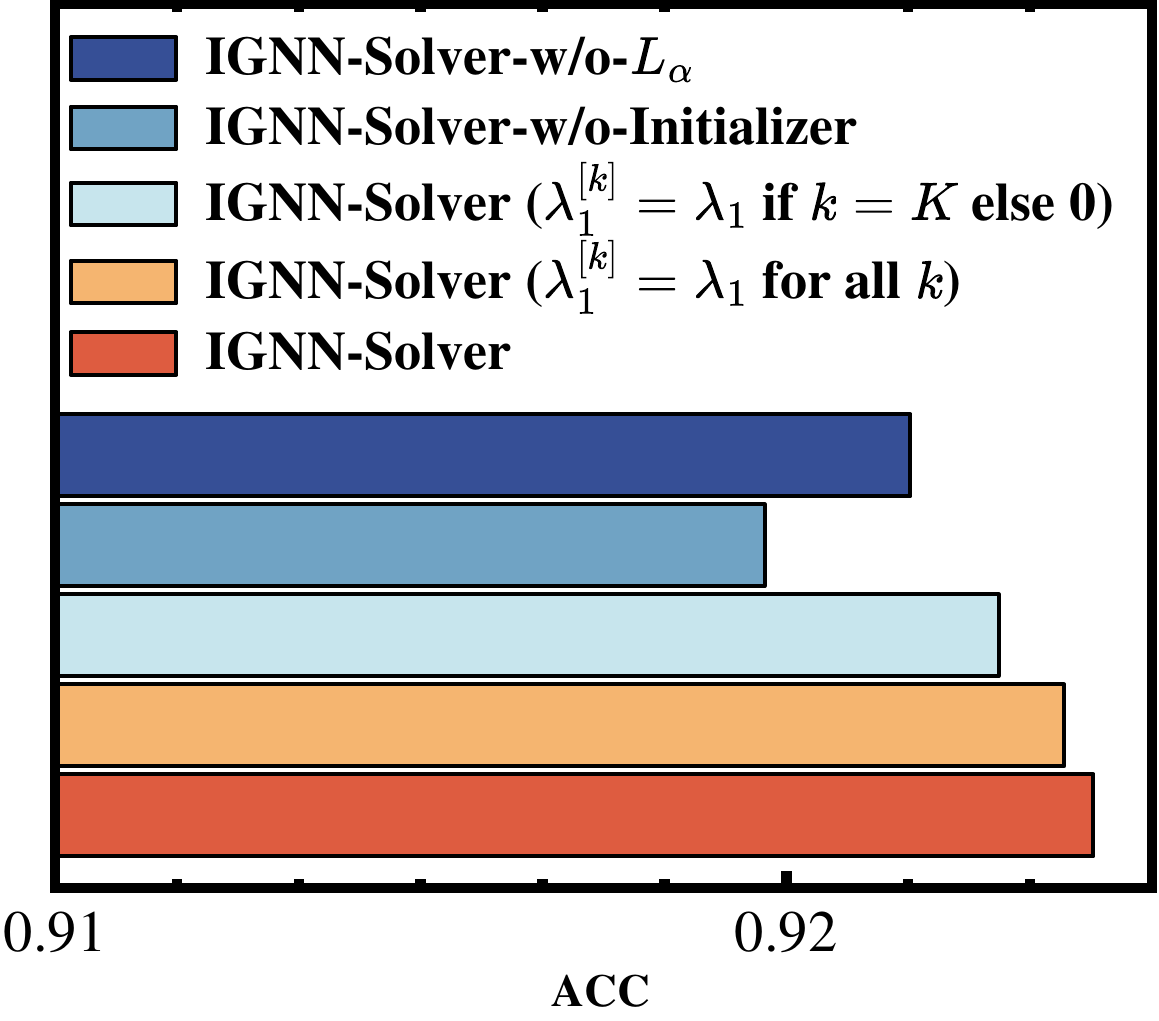}
        \caption{Reddit}
    \end{subfigure}
    \begin{subfigure}{0.3\textwidth}
        \includegraphics[height=4.5cm]{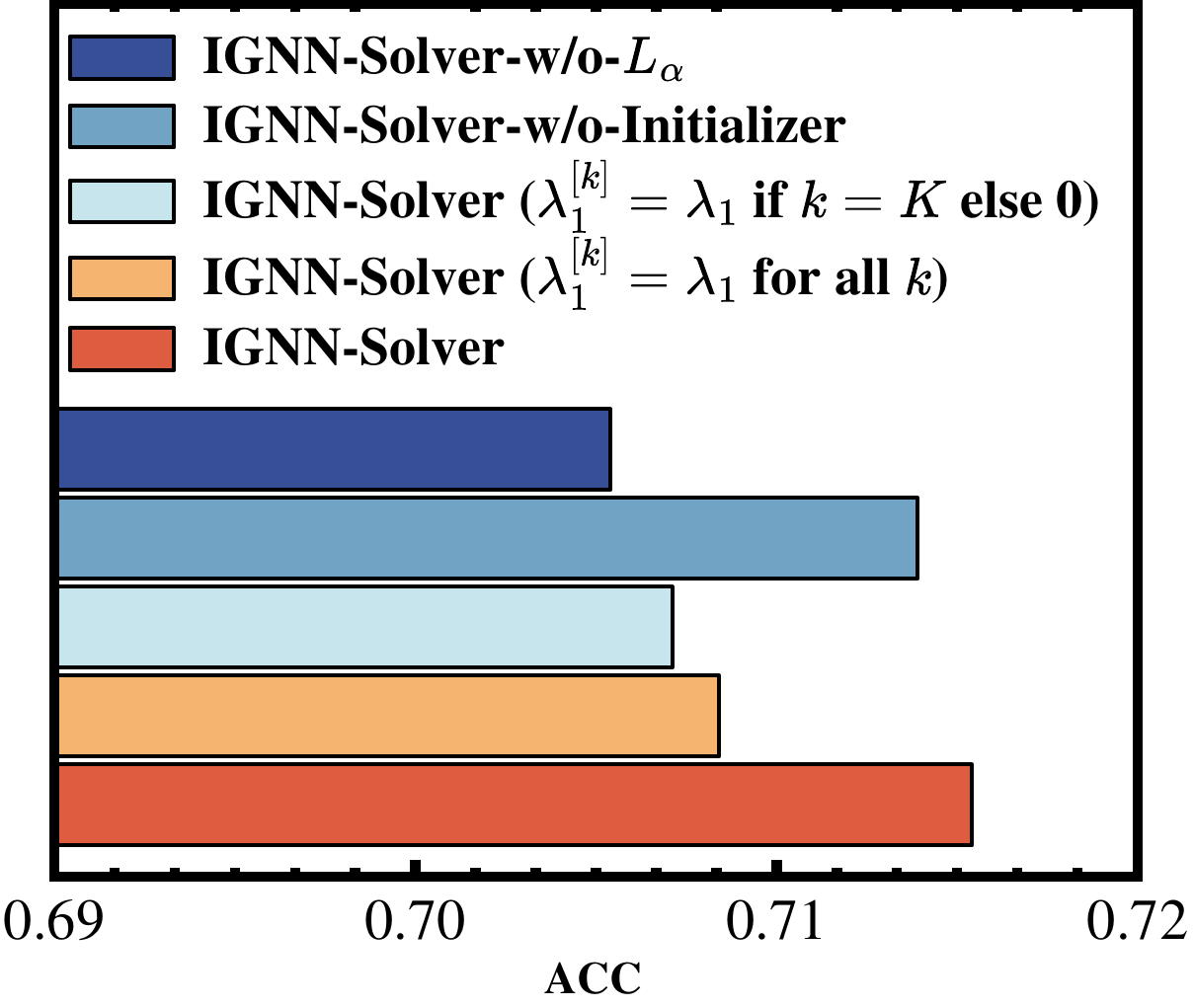}
        \caption{obgn-arxiv}
    \end{subfigure}
    \begin{subfigure}{0.3\textwidth}
        \includegraphics[height=4.5cm]{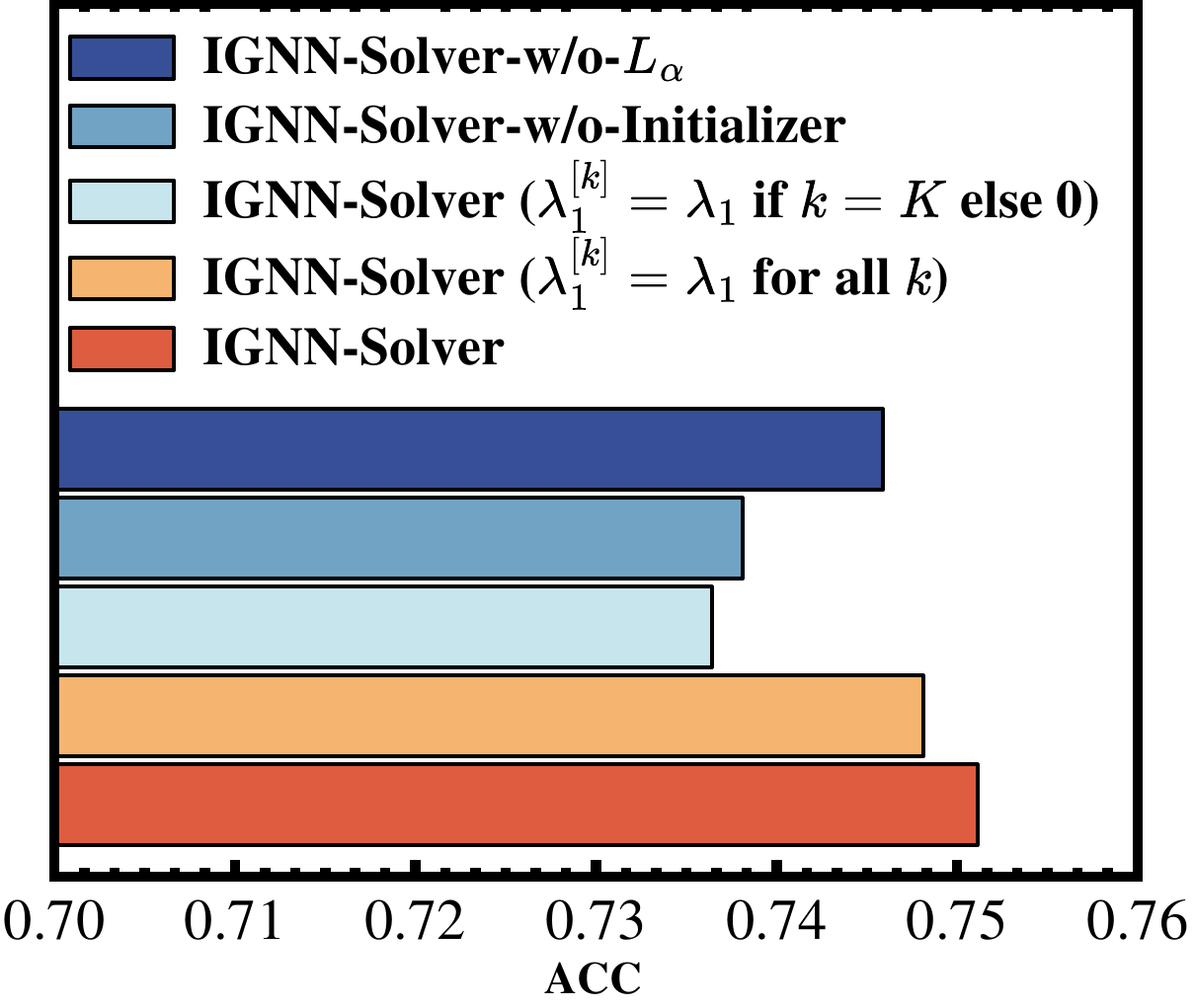}
        \caption{obgn-products}
    \end{subfigure}
    \begin{subfigure}{0.3\textwidth}
        \includegraphics[height=4.5cm]{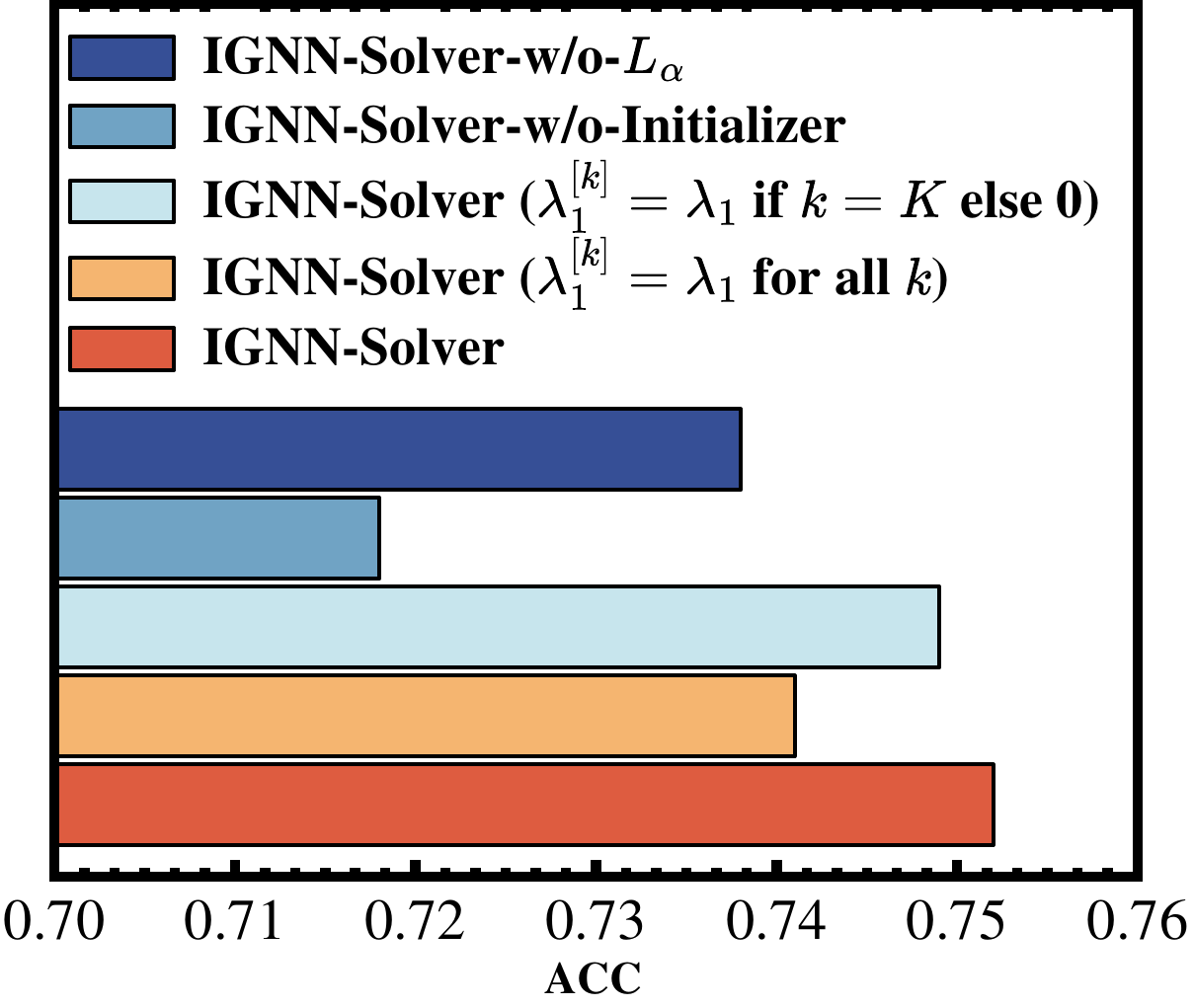}
        \caption{Citeseer}
    \end{subfigure}
    \begin{subfigure}{0.3\textwidth}
        \includegraphics[height=4.5cm]{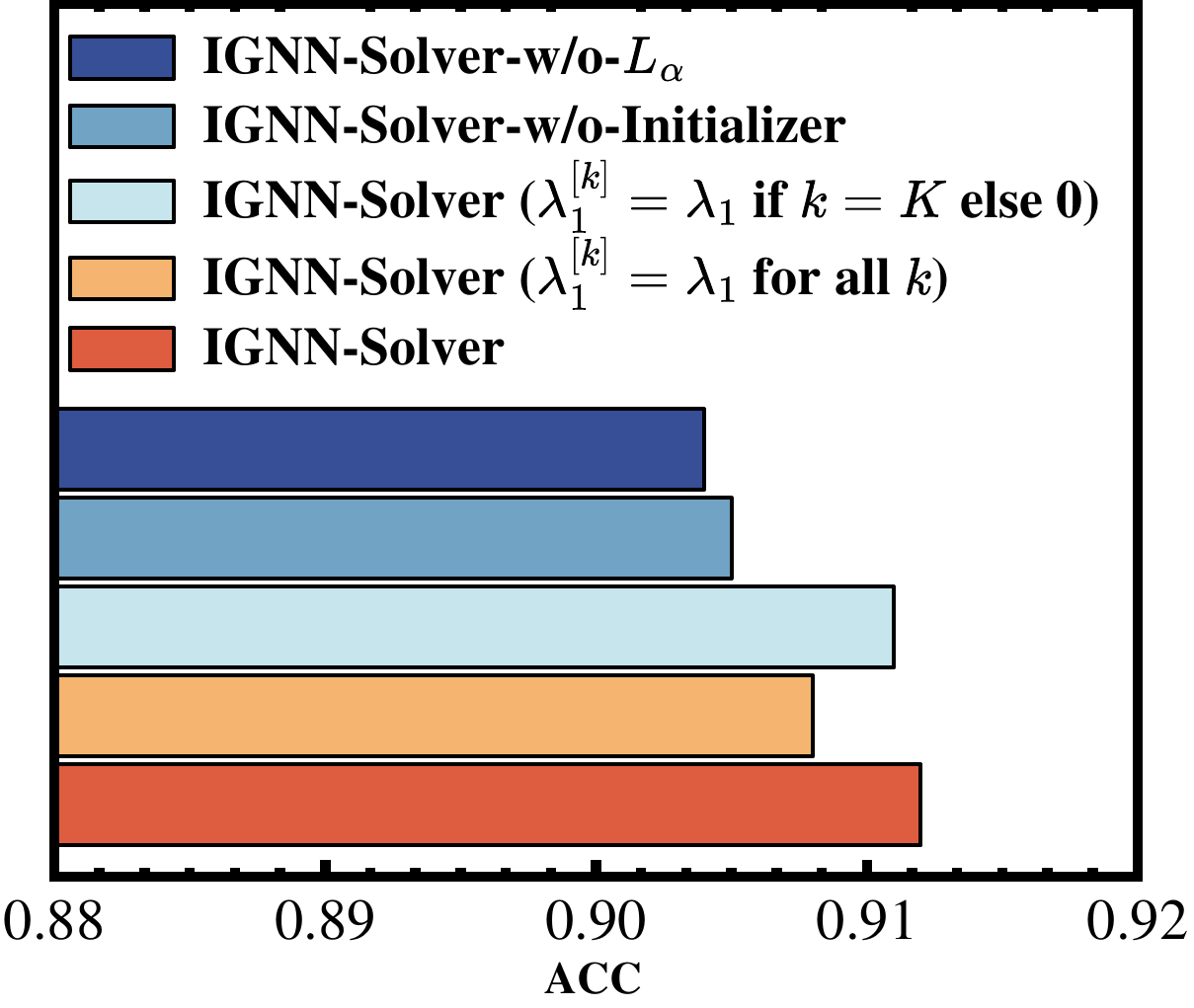}
        \caption{ACM}
    \end{subfigure}
    \begin{subfigure}{0.3\textwidth}
        \includegraphics[height=4.5cm]{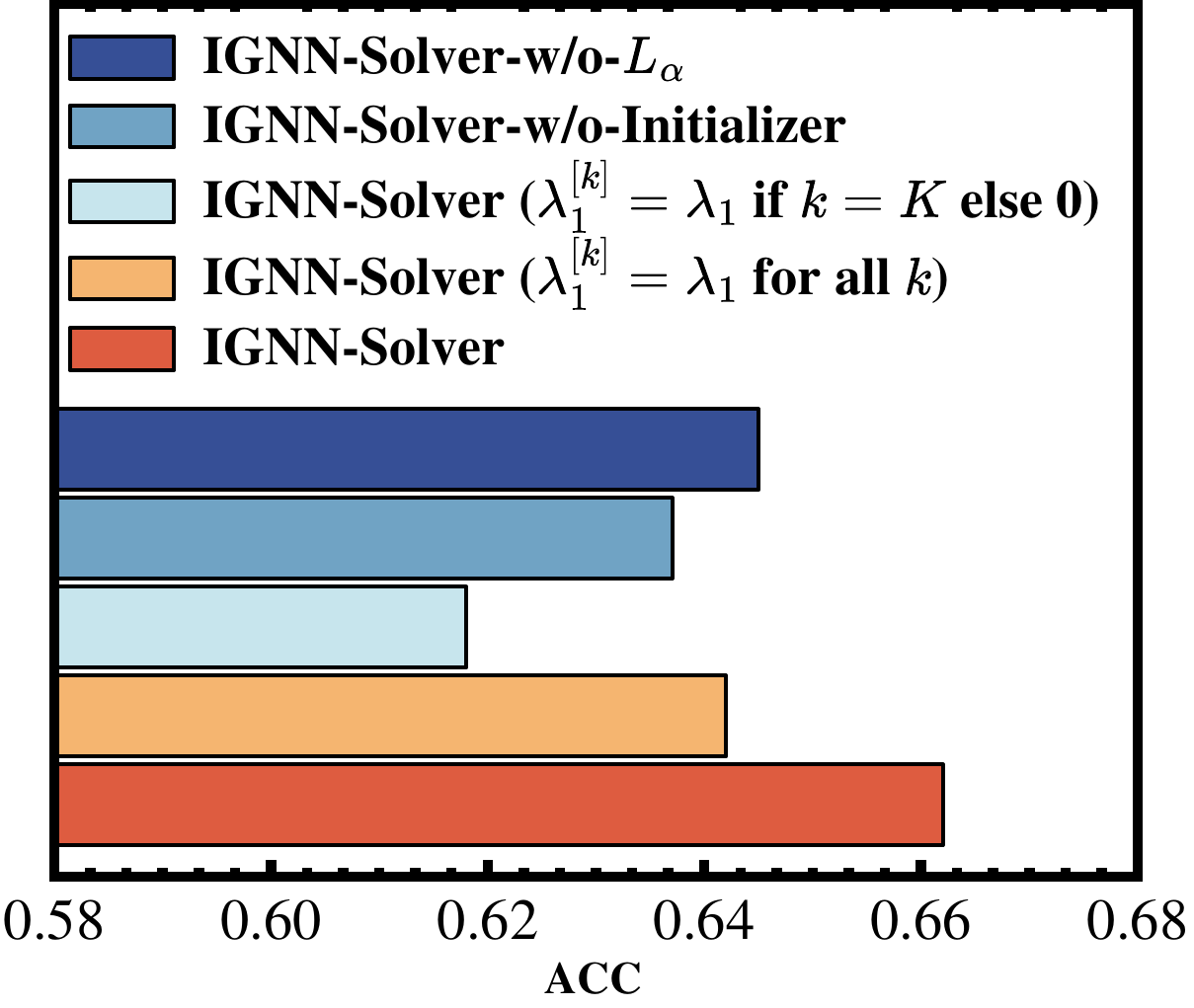}
        \caption{CoraFull}
    \end{subfigure}
    \begin{subfigure}{0.3\textwidth}
        \includegraphics[height=4.5cm]{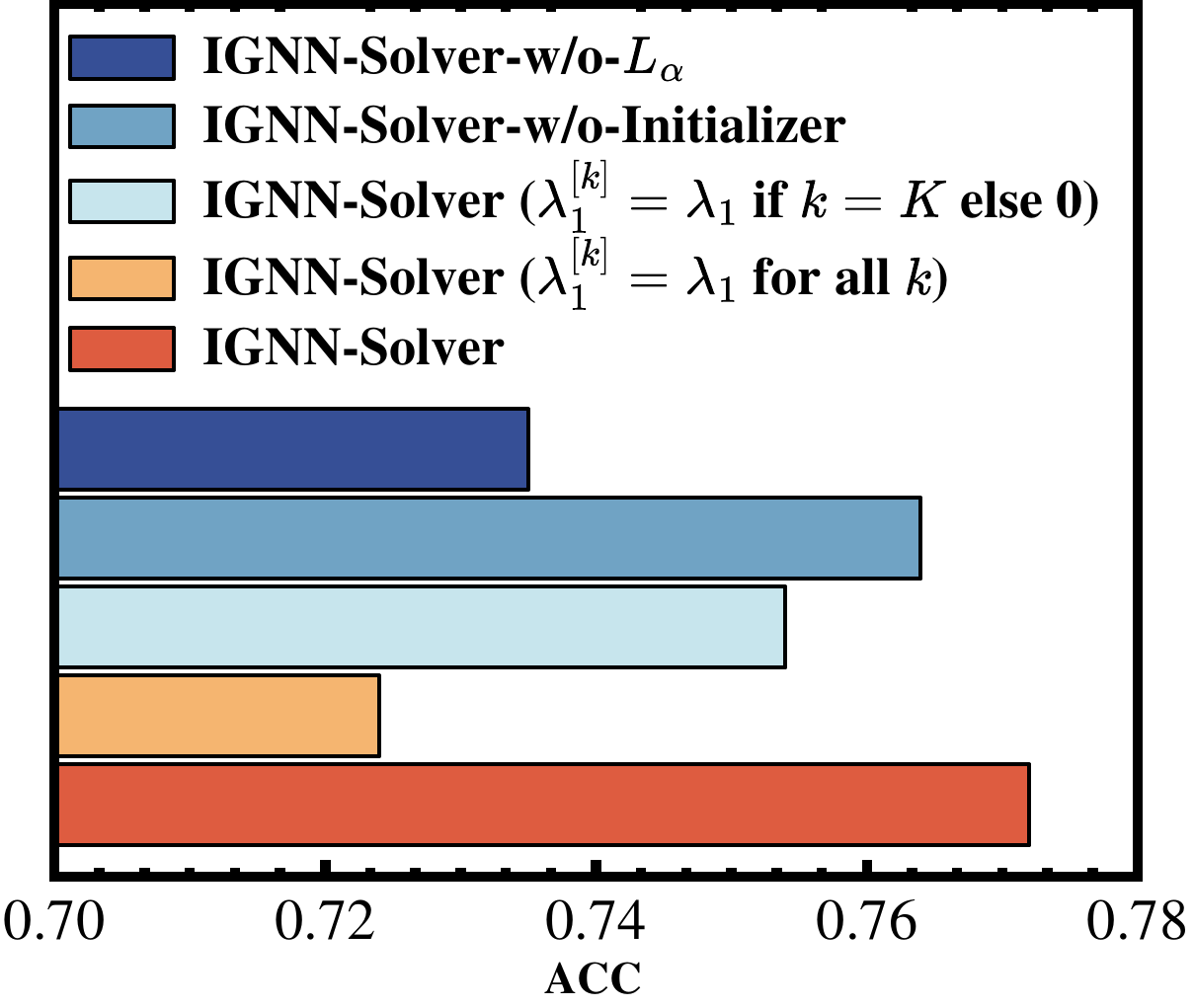}
        \caption{BlogCatalog}
    \end{subfigure}
    \begin{subfigure}{0.3\textwidth}
        \includegraphics[height=4.5cm]{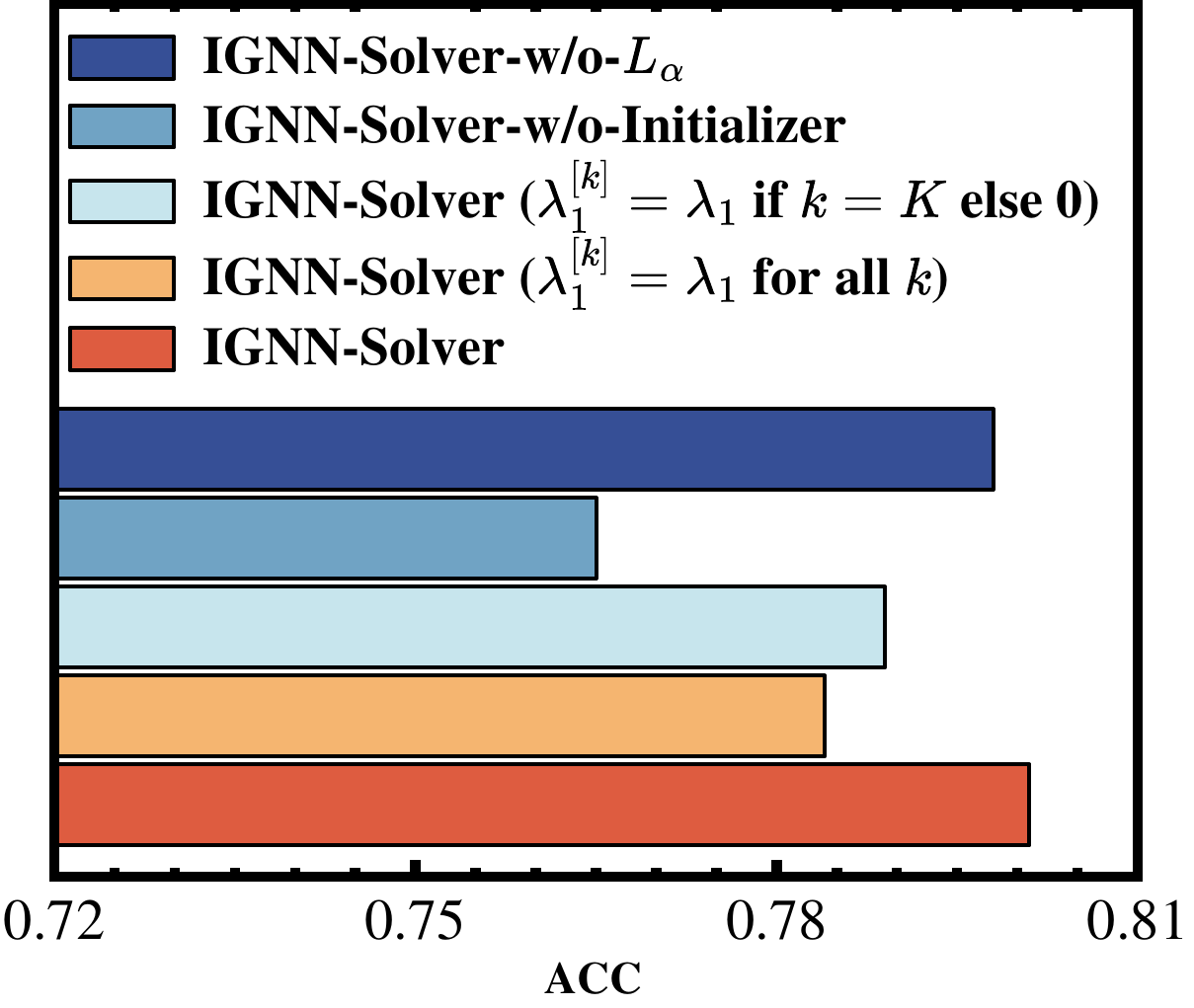}
        \caption{Flickr}
    \end{subfigure}
    \caption{Ablation studies for IGNN-Solver. This study shows that all components contribute significantly to accelerating the convergence process.}
    \label{F_Ablation}
\end{figure*}

\subsubsection{Ablation study}\label{sec_Ablation}
In this section, we analyze the effect of different loss components on the IGNN-Solver. The performance of IGNN-Solver and its variants on the Citeseer dataset is shown in Figure~\ref{F_Ablation}, with similar patterns observed across other datasets and solvers.
For the main convergence loss $\mathcal{L}_{\text{conv}}$ in fixed-point iterations \eqref{eq10}, we design two contrasting schemes concerning its weight $\lambda_1^{[k]}$s: setting them to a constant value (i.e., $\lambda_1^{[k]} = \lambda_1 $ for all $k$), or setting all values except the $K$-th term to zero (i.e., $\lambda_1^{[k]} = \lambda_1$ if $k = K$ else $0$). Both variant solvers still perform well, yet our suggested monotonically increasing scheme (i.e., emphasizing the later steps to a greater extent) exhibits the best performance.

Furthermore, removing either the initializer or the alpha loss $\mathcal{L}_\alpha$ negatively impacts performance, with the former having a more detrimental impact on the solver. Nonetheless, all these ablation configurations significantly outperform methods without a solver, underscoring the advantages of employing a custom learnable solver for the IGNN model.

\section{Conclusion}
\label{sec6}
This paper introduces IGNN-Solver, a novel graph neural solver tailored for achieving fast fixed-point solving in IGNNs. 
Unlike previous approaches that impose constraints on the structure or parameters of the implicit layer, we propose to leverage the important idea that implicit models separate the representation from the forward computation.
By leveraging the generalized Anderson Acceleration method and parameterizing it with a tiny GNN, IGNN-Solver learns iterative updates as a graph-dependent temporal process. To address large-scale graph tasks, we further tailored sparsification and storage compression methods for the IGNN solver and integrated them into its design. Our extensive experiments demonstrate the significant acceleration in inference achieved by IGNN-Solvers, with a speedup ranging from $1.5\times$ to $8\times$, all while maintaining accuracy. Notably, this acceleration is particularly pronounced as the scale of the graph increases. These findings underscore the potential of IGNN-Solver for large-scale end-to-end deployment in real-world applications.

Looking ahead, IGNN-Solver opens new avenues for the integration of learnable solvers in implicit GNNs, bridging the gap between solver efficiency and adaptability to diverse graph structures. Future research could explore extending IGNN-Solver to handle dynamic or heterogeneous graphs, further broadening its applicability. Additionally, investigating the interplay between solver generalization and graph topology holds promise for uncovering deeper insights into the dynamics of graph-based learning. With its ability to combine speed, scalability, and accuracy, IGNN-Solver represents a pivotal step toward making implicit graph models a viable choice for large-scale real-world systems. We hope that IGNN-Solver inspires further innovations in learnable solvers for IGNNs, driving progress in scalable and efficient algorithms on graphs.



\bibliographystyle{IEEEtran}
\bibliography{IEEEtran}

\twocolumn[
\begin{center}
    {\Large \textbf{Supplementary Material}}\\ 
    \vskip 0.1in \textbf{IGNN-Solver: A Graph Neural Solver for Implicit Graph Neural Networks}
\end{center}
\vspace{1em}
]

\appendices

\section{Experimental Details}\label{secExperimental_Details}
The experiments are conducted on nine public datasets of different scales for node classification tasks, including five widely adopted benchmark datasets, Citeseer, ACM, CoraFull, BlogCatalog, Flickr, and four large-scale benchmark datasets, Amazon-all, Reddit, ogbn-arxiv and ogbn-products; as well as five bioinformatics benchmarks for graph classification tasks: MUTAG, PTC\_MR, COX2, PROTEINS, NCI1. 
We endeavor to conduct inference testing under almost identical hyperparameter conditions as previous work \cite{MIGNN, EIGNN, IGNN, AMGCN, GCN, GAT, SGC, APPNP, JKNet}, including performance and efficiency comparisons.
All the Experiments are conducted independently five times (i.e., using five random seeds) on a machine with Intel(R) Xeon(R) Gold 6138 CPU @ 2.00GHz with a single 3090 GPU.
A detailed description of all tasks and additional details are provided below.

\subsection{Datasets}
\begin{itemize}
	\item\textbf{{Citeseer}}~\cite{GCN}: This dataset is a citation network of research papers, which are divided into six categories. The citation network takes 3,327 scientific papers as nodes and 4,732 citation links as edges. The feature of each node in the dataset is a word vector to describe whether the paper has the corresponding words or not.
	\item\textbf{{ACM}}~\cite{yun2019graph}:
	It is a citation network dataset, where nodes represent papers and node features are constructed by the keywords. Papers are divided into three categories according to their types of conferences.
	\item\textbf{{CoraFull}}~\cite{bojchevski2018deep}:
	Similar to the Citeseer dataset, CoraFull is a well-known citation network labeled based on the paper topic, which contains 19,793 scientific publications. CoraFull is classified into one of 70 categories, where nodes represent papers and the edges represent citations.
	\item\textbf{{BlogCatalog}}~\cite{meng2019co}:
	This dataset is a social relationship network. The graph is composed of bloggers and their social relationships (such as friends). Node attributes are constructed by keywords in the user profile. The labels represent bloggers' interests. All nodes are divided into six categories.
	\item\textbf{{Flickr}}~\cite{demonet}:
	It is a graphic social network where nodes represent users and edges correspond to the friendships among users. All the nodes are divided into nine classes according to the interest groups of users.
	\item\textbf{{Amazon-all}}~\cite{Amazonall}:
	The widely-used benchmark dataset Amazon-all encompasses the Amazon product co-purchasing network dataset. This dataset represents products as nodes and co-purchases as edges. It includes 58 product types, each with over 5,000 products, selected from a total pool of 75,149 product types.
	\item\textbf{{Reddit}}~\cite{Reddit}: The Reddit dataset is a social network where nodes represent posts, and edges indicate that the same user commented on two connected posts. Each node contains 602 dimensional features.
	\item\textbf{{ogbn-arxiv}}~\cite{OGB}: The ogbn-arxiv dataset is a citation network between all Computer Science (CS) arXiv papers. Each node is an arXiv paper, and each directed edge indicates that one paper cites another. Each paper comes with a 128-dimensional feature vector obtained by averaging the embeddings of words in its title and abstract. The task is to predict the 40 subject areas of arXiv CS papers.
	\item\textbf{{ogbn-products}}~\cite{OGB}: The ogbn-products dataset contains an undirected and unweighted graph, representing an Amazon product co-purchasing network. Nodes represent products sold on Amazon, and edges between two products indicate that the products are purchased together. The task is to predict the category of a product in a multi-class classification setup, where the 47 top-level categories are used for target labels.
	\item \textbf{MUTAG}~\cite{yanardag2015deep}: A dataset of nitroaromatic compounds, where the task is to predict the mutagenic effect on a bacterium. It contains 188 graphs with an average of 17.9 nodes per graph.
	\item \textbf{PTC-MR}~\cite{yanardag2015deep}: A dataset of 344 chemical compounds that report carcinogenicity for male and female rats. The dataset is used to predict chemical carcinogenicity, where each compound is represented as a graph with atoms as nodes and chemical bonds as edges. The average number of nodes per graph is 14.3, and the task is binary classification to determine whether a compound is carcinogenic or not. The node features represent atom types and other chemical properties.
	\item \textbf{COX2}~\cite{yanardag2015deep}: A dataset of 467 cyclooxygenase-2 (COX-2) inhibitors. The dataset contains chemical compounds that inhibit COX-2 enzyme activity, which is important in inflammation and pain. Each compound is represented as a graph where atoms are nodes, and chemical bonds are edges.
	\item \textbf{PROTEINS}~\cite{yanardag2015deep}: This dataset consists of proteins represented as graphs, where nodes are secondary structure elements, and edges indicate neighborhood in the amino-acid sequence or in 3D space. It includes 1,113 graphs with an average of 39.1 nodes per graph.
	\item \textbf{NCI1}~\cite{yanardag2015deep}: A dataset of chemical compounds screened for activity against non-small cell lung cancer and ovarian cancer cell lines. It contains 4,110 graphs with an average of 29.8 nodes per graph.
\end{itemize}

\begin{table*}
	\caption{Hyper-paramaters setting on training. *In addition, we linearly increase the loss weight $\lambda_1$ from $0$ to its set value throughout all IGNN-Solver training steps.}
	\label{Hp_setting}
	\centering
	\resizebox{0.9\textwidth}{!}{
		\begin{tabular}{clllllllllll}
			\toprule
			Scale                  & Datasets      & nhid & dropout & lr    & $K$ & $\text{epoch}_\text{max}$ & training & test      & $\lambda_1$* & $\lambda_2$ & $\lambda_3$
			\\ \midrule
			\multirow{5}{*}{Small} & Citeseer      & 128  & 0.5     & 0.002 & 10  & 100                       & 360      & 1000      & 0.1          & 5           & 1e-4        \\
			& ACM           & 128  & 0.5     & 0.001 & 10  & 100                       & 180      & 1000      & 0.3          & 5           & 1e-4        \\
			& CoraFull      & 512  & 0.5     & 0.002 & 15  & 300                       & 4200     & 1000      & 0.3          & 5           & 1e-4        \\
			& BlogCatalog   & 512  & 0.5     & 0.001 & 10  & 100                       & 360      & 1000      & 0.5          & 5           & 1e-4        \\
			& Flickr        & 512  & 0.5     & 0.002 & 15  & 200                       & 540      & 1000      & 0.1          & 5           & 1e-5        \\
			\midrule
			\multirow{4}{*}{Large} & Amazon-all    & 128  & 0.5     & 0.005 & 20  & 1000                      & 16970    & 28285     & 0.3          & 5           & 1e-5        \\
			& Reddit        & 512  & 0.5     & 0.005 & 15  & 2000                      & 139,779  & 46,593    & 0.5          & 5           & 1e-5        \\
			& ogbn-arxiv    & 128  & 0.5     & 0.001 & 20  & 2000                      & 90,941   & 48,603    & 0.5          & 5           & 1e-5        \\
			& ogbn-products & 128  & 0.5     & 0.001 & 20  & 2000                      & 196,615  & 2,213,091 & 0.5          & 5           & 1e-5        \\
			\bottomrule
		\end{tabular}
	}
\end{table*}

\subsection{Hyper-parameters Setting in IGNN-Solver}
\label{sec_Hp_setting}
We present in Table \ref{Hp_setting} all the hyper-parameters preset by IGNN-Solver across all datasets,
where $K$ represents the threshold of maximum iteration in IGNN-Solver, $\text{epoch}_\text{max}$ means the maximum training steps in IGNN.
It is worth noting that we set the value of $\lambda_1$ in a linearly increasing manner to penalize the intermediate estimation error when the training step $K$ is large. Similarly, we set the value of $\lambda_3$ in a linearly decreasing manner to prevent overfitting $\alpha$ during later training epochs.

\section{Training Strategies}\label{Appendix_Training_Strategies}
\subsection{IGNN Training with frozen IGNN-Solver}
During the training processing of the IGNN model, we assume that both the initializer $h_\phi$ and the neural solver $s_{\xi}$ of IGNN-Solver have approached the desired state via the training method outlined in Section~\ref{sec4}, and they will no longer be updated in the subsequent model training.
For a given IGNN layer $f_\theta$, as well as the input graph $\mA$ and feature matrix $\mX$, we define:
\begin{equation}\label{14}
	g_\theta(\mZ^{\star}, \widehat{\mA}, \mX) := f_\theta(\mZ^{\star}, \widehat{\mA}, \mX)-\mZ^{\star}=0,
\end{equation}
and utilize the IGNN-Solver to solve \eqref{14}, obtaining the fixed point $\mZ^\star$.
Furthermore, the fixed point $\mZ^\star$ is passed through a linear layer to obtain an embedding $\mH = \text{Linear}(\mZ^\star)$. Then we, jointly with downstream tasks, compute the Binary Cross Entropy (BCE) loss~\cite{BCE}
to back-propagate and update the parameters of $f_\theta$ in a supervised way. The complete process is illustrated in Algorithm~\ref{Algorithm_3}.

\begin{algorithm}
    \renewcommand{\algorithmicrequire}{\textbf{Input:}}
    \renewcommand{\algorithmicensure}{\textbf{Output:}}
    \caption{IGNN model training}
    \label{Algorithm_3}
    \begin{algorithmic}[1]
        \Require  graph matrix $\widehat{\mA}$, feature matrix $\mX$, fixed IGNN-Solver $s_{\xi}$ and initializer $h_\phi$
        \State Initialize $\mZ^{[0]} = h_\phi(\mX)$ and randomly initialize $f_\theta$
    
        \While{Stopping condition is not met}
    
        \State $\mZ^{\star} \gets \text{Solve}  \ f_\theta(\mZ^{\star}, \widehat{\mA}, \mX)-\mZ^{\star}=0 $
        \Comment{via frozen $s_{\xi}$, $\widehat{\mA}$ and initial value $\mZ^{[0]}$}
    
        \State $\mH \gets \text{Linear}(\mZ^\star)$
    
        \State $\mathcal{L}_\text{task} \gets \text{BCE} (  \mH , \mY   ) $
        \Comment{computing the label loss}

        \State Back-propagate $\mathcal{L}_\text{task}$ to update $f_\theta$
        \EndWhile
        \State \Return $f_\theta$
    \end{algorithmic}
\end{algorithm}

\subsection{Advanced IGNN Training Strategy with IGNN-Solver}
Given the fact that model parameter $\theta$ gradually updates only with the model training iterations (i.e., we assume $
\left\|\theta_{t+1}-\theta_{t}\right\|
$ is only related to $\theta_{t}$ at training step $t$), we propose training the lightweight IGNN-Solver $s_{\xi}, h_{\phi}$ and the large model $f_\theta$ in an alternating way. Specifically, we adopt the following procedure:

\begin{enumerate}[label=(\roman*)]
	\item Warm up and train the IGNN model and its solver (IGNN-Solver) for some steps.
	\item Fix the IGNN-Solver $s_{\xi}, h_{\phi}$ and solving the fixed points of $f_\theta$ in IGNN via Algorithm \ref{Algorithm_3} for $T_1$ steps. \label{step_2}
	\item Fix the current model parameters $\theta$ and start fine-tuning the IGNN-Solver $s_{\xi}, h_{\phi}$ via Algorithm \ref{Algorithm_2} over some $T_2$ steps. \label{step_3}
	\item Repeat steps \ref{step_2} and \ref{step_3} until reaching the maximum training steps for the IGNN model.
\end{enumerate}

Here $T_1$ is approximately 4\% to 10\% of $T_2$, adjusted as needed. The sum of all $T_2$ values is referred to as $\text{epoch}_\text{max}$.
It is reassuring that the additional cost of fine-tuning the IGNN-Solver (step \ref{step_3}) is sufficiently offset by the substantial benefits of accelerated solving (step \ref{step_2}).
Moreover, we are surprised to find that the proposed IGNN-Solver does not exhibit a decline in expressive capability for excessively small or large $T_1$ values. This indicates that thanks to its robust stability, accidentally setting $T_1$ high does not significantly affect the normal training of the IGNN. Conversely, if $T_1$ is set slightly lower, the model still demonstrates good generalization capabilities. Thus, the IGNN-Solver shows low sensitivity to $T_1$.

\section{More Experiments on IGNN-Solver}
\subsection{Training Dynamics}
\begin{figure}
    \centering
    \includegraphics[width=0.9\linewidth]{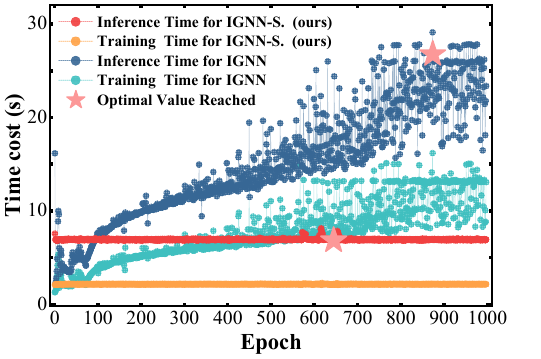}
    \caption{Training time and inference time(s) per epoch of IGNN and IGNN-Solver on Amazon-all datasets.
    }
    \label{F_Visualization}
\end{figure}

In this section, we present the time cost trends of IGNN and IGNN-Solver during training and inference on the Amazon dataset. From Figure~\ref{F_Visualization}, it is evident that IGNN starts relatively fast overall at the beginning of training. However, as the training of IGNN progresses, the model becomes increasingly complex, leading to a sharp rise in the time cost for fixed-point computation. After 100 epochs, this cost remains persistently high and fluctuates continuously. This issue arises because the IGNN reaches the maximum number of iterations without achieving the preset minimum error.

On the contrary, our IGNN-Solver demonstrates significantly lower training and inference time consumption compared to the former and maintains a stable state throughout the process. This is attributed to the solver's consistent memory consumption and its rapid fixed-point computation capability, which drives the unique advantage of IGNN-Solver.

\subsection{Long-Range Dependencies (LRD)}
To evaluate the capability of GNNs in capturing long-range dependencies within graphs, we follow the methodology presented in \cite{IGNN} by creating a synthetic dataset called Chains. In this dataset, the objective is to classify nodes positioned in a chain of length l. The class information is provided sparsely, appearing only as a feature in one end node of the chain. Our experimental setup consists of small datasets: a training set with 20 nodes, a validation set with 100 nodes, and a test set with 200 nodes. For simplicity, we focus on binary classification. Following the experimental protocol outlined in IGNN, we implement and evaluate four representative baseline models.

\begin{figure}
    \centering
    \includegraphics[width=0.75\linewidth]{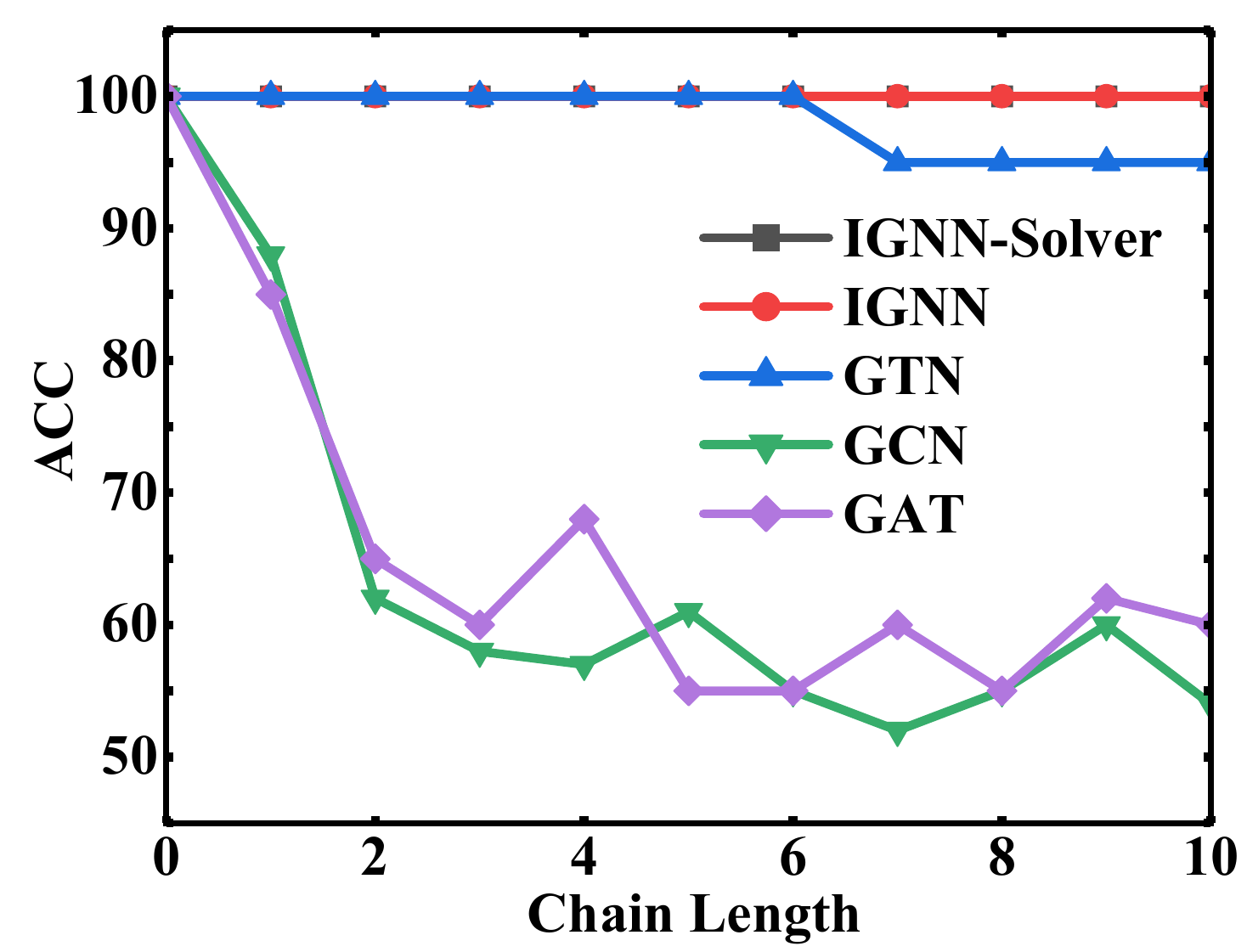}
    \caption{Average accuracy with respect to the length of chains.}
    \label{fig:LRD}
\end{figure}

As demonstrated in Figure~\ref{fig:LRD}, both IGNN-Solver and IGNN effectively capture long-range dependencies on longer chains, whereas the performance of the graph transformer (GTN)~\cite{yun2019graph} slightly decreases when the chain length exceeds $6$. On the other hand, convolutional GNNs with a fixed number of iterations $(T = 2)$, like GCN and GAT, face challenges in making meaningful predictions as the chain length grows due to their limited capacity to capture dependencies beyond two hops. Interestingly, increasing the number of iterations $(T)$ for these models does not lead to significant improvement in this case.

\vfill

\end{document}